\documentclass[journal]{IEEEtran}
\usepackage{graphics} 
\usepackage{epsfig} 
\usepackage{times} 
\usepackage{amsmath} 
\usepackage{amssymb}  
\usepackage{cite}
\usepackage{epstopdf}
\usepackage{subfigure,color,balance}
\usepackage{verbatim}
\usepackage{cases}
\usepackage{enumerate}
\usepackage{booktabs}
\usepackage{balance}
\usepackage{mathbbol}
\usepackage{algorithm}
\usepackage{algorithmicx}
\usepackage{algpseudocode}
\usepackage{comment}
\usepackage{gensymb}
\usepackage{textcomp}
\usepackage{threeparttable}

\usepackage{tikz}
\newcommand*{\circled}[1]{\lower.7ex\hbox{\tikz\draw (0pt, 0pt)%
		circle (.5em) node {\makebox[1em][c]{\small #1}};}}
\usepackage[colorlinks=true,      
linkcolor=black,      
citecolor=black,      
filecolor=black,      
urlcolor=blue]{hyperref}

\newtheorem{remark}{Remark}

\def\0{{\bf 0}}
\def\1{{\bf 1}}

\def\eg{{\em e.g.}}
\def\ie{{\em i.e.}}

\begin{document}
%
\title{
Mixed Cloud Control Testbed: \\Validating Vehicle-Road-Cloud Integration via Mixed Digital Twin
}

\author{Jianghong Dong, Qing Xu, Jiawei Wang,  Chunying Yang, Mengchi Cai, Chaoyi Chen,\\ Jianqiang Wang,  and Keqiang Li 
	\thanks{This work is supported by National Key R\&D Program of China with 2021YFB1600402, Tsinghua University-Didi Joint Research Center for Future Mobility, China Intelligent and Connected Vehicles (Beijing) Research Institute Co., Ltd. and Dongfeng Automobile Co., Ltd. Corresponding author: Jiawei Wang.}
	\thanks{Jianghong Dong, Qing Xu, Jiawei Wang, Mengchi Cai, Chaoyi Chen, Jianqiang Wang and Keqiang Li are with the School of Vehicle and Mobility, Tsinghua University, Beijing, China, and with Tsinghua University-Didi Joint Research Center for Future Mobility, Beijing, China.(\{djh20,wang-jw18,cmc18,chency19\}@mails.tsinghua.edu.cn,  \{qingxu,wjqlws,likq\}@tsinghua.edu.cn) }
	\thanks{Chunying Yang was with the School of Vehicle and Mobility, Tsinghua University, and with the School of Transportation Science and Engineering, Beihang University. She is now with Beijing Idriverplus Technology Co., Ltd. (yangchunying@idriverplus.com)}%
	%
}

\maketitle

\begin{abstract}
	
Reliable and efficient validation technologies are critical for the recent development of multi-vehicle cooperation and vehicle-road-cloud integration. In this paper, we introduce our miniature experimental platform, Mixed Cloud Control Testbed (MCCT), developed based on a new notion of Mixed Digital Twin (mixedDT). Combining Mixed Reality with Digital Twin, mixedDT integrates the virtual and physical spaces into a mixed one, where physical entities coexist and interact with virtual entities via their digital counterparts. Under the framework of mixedDT, MCCT contains three major experimental platforms in the physical, virtual and mixed spaces respectively, and provides a unified access for various human-machine interfaces and external devices such as driving simulators. A cloud unit, where the mixed experimental platform is deployed, is responsible for fusing multi-platform information and assigning control instructions, contributing to synchronous operation and real-time cross-platform interaction. Particularly, MCCT allows for multi-vehicle coordination composed of different multi-source vehicles (\eg, physical vehicles, virtual vehicles and human-driven vehicles). Validations on vehicle platooning demonstrate the flexibility and scalability of MCCT. 


\end{abstract}


\section{Introduction}

\IEEEPARstart{R}{ecent}
advances in vehicle-to-everything (V2X) network and 5G communication have allowed for data transmission with high reliability and low latency between connected
and automated vehicles (CAVs), roadside facilities, and cloud servers. The notion of vehicle-road-cloud integration or cloud control has subsequently attracted extensive attention and promised transportation revolution~\cite{xia2015cloud,li2020Principles,chu2021cloud}. Typical research include multi-vehicle platoon control~\cite{li2022cooperative}, cooperation for multiple connected vehicles at unsignalized intersections~\cite{xu2018distributed}, and  urban traffic network management~\cite{liu2021reservation}. These research demonstrate the potential of vehicle-road-cloud integration in significantly reducing energy consumption and improving traffic efficiency~\cite{guanetti2018control,xia2019cloud}.

To validate the aforementioned potential of multi-vehicle collaboration and vehicle-road-cloud integration, existing research mostly focus on the tool of traffic simulations; see, \emph{e.g.},~\cite{li2022cooperative,xu2018distributed,liu2021reservation}. However, many practical factors in the vehicle-road-cloud integration system, which is essentially a complex Cyber Physical System~\cite{jia2015survey}, cannot be accurately replicated in the simulations. Typical factors include vehicle dynamics, human driving behaviors, uncertainties in sensor measurement, and delays in communication and computation process. On the other hand, field experiments via deploying real-world CAVs and roadside facilities take great labor and material costs, and are extremely time- and space-inefficient~\cite{feng2021intelligent}. One representative real-world experimental platform is Mcity~\cite{Mcity-website}, which mainly focuses on the verification of autonomous driving algorithms on urban roads~\cite{xu2021system}. Given the limited and fixed road infrastructures in Mcity, it is still intractable to conduct large-scale experiments and adjust the experimental environment~\cite{feng2020safety}. 
As for multi-vehicle collaboration, despite the existence of a few field tests for small-scale particular traffic scenarios (see, \emph{e.g.},~\cite{stern2018dissipation,liao2021cooperative}), it is still non-trivial for comprehensive demonstrations on large-scale vehicle-road-cloud integration. 

To address these problems, miniature experimental platforms~\cite{hoffmann2004stanford,how2008real,paull2017duckietown}, also known as scaled platforms or sand tables, promise to fill the gap between numerical simulations and field experiments. Miniature platforms are established in a real-world physical environment, where miniature vehicles and roadside facilities are deployed. They provide hardware and software environments with closer operating features to practical application scenarios compared with numerical simulations. Unlike field experiments, such platforms are also believed to be much more convenient for further adjustment, reproduction or migration at affordable costs, according to various traffic scenarios and demands~\cite{stager2018scaled}. 
Given the aforementioned benefits, we establish our miniature experimental platform Mixed Cloud Control Testbed (MCCT), which 
allows for real-world demonstrations and various research on vehicle-road-cloud integration.

\subsection{Literature Review}
Given the aforementioned benefits of miniature experiments, there have already existed several attempts in the academy for building miniature CAV  validation platforms. Precisely, existing  platforms can be classified into three categories, according to different application levels: 1) Single-vehicle autonomous driving, including the platforms for verifying basic autonomous driving technologies on urban roads~\cite{karaman2017project,goldfain2019autorally,o2019f1} and those focusing on more complex 
environments, such as real-time control in rough 3D terrain~\cite{keivan2013realtime} and drift control~\cite{gonzales2016autonomous}.
2) Multi-vehicle coordination control. These platforms are mostly developed with the aim of validating cooperation algorithms at one particular traffic scenario, including on/off ramps~\cite{stager2018scaled}, multi-lane highways~\cite{hyldmar2019fleet}, and non-signalized intersections~\cite{fok2012platform}. 3) Vehicle-road collaboration or vehicle-cloud collaboration. Two representative platforms for this level are~\cite{wan2011general}, which tests navigation for individual vehicles via the roadside units, and~\cite{sasaki2016vehicle}, which demonstrates the potential of utilizing edge computing to directly issue control commands to CAVs. 
To the best of our knowledge, there is currently no miniature platform that supports the comprehensive research on vehicle-road-cloud integration, which covers the aforementioned three application levels. Furthermore, considering that there could still be a lot of human-driven vehicles (HDVs) on the road in the near future~\cite{zheng2020smoothing,stern2018dissipation}, it remains an open question to incorporate human drivers into the miniature platform whilst preserving a real-world driving behavior~\cite{la2012development}.

\begin{figure}[t]
	\vspace{1mm}
	\centering
	\subfigure[Classical DT]
	{\includegraphics[scale=0.55]{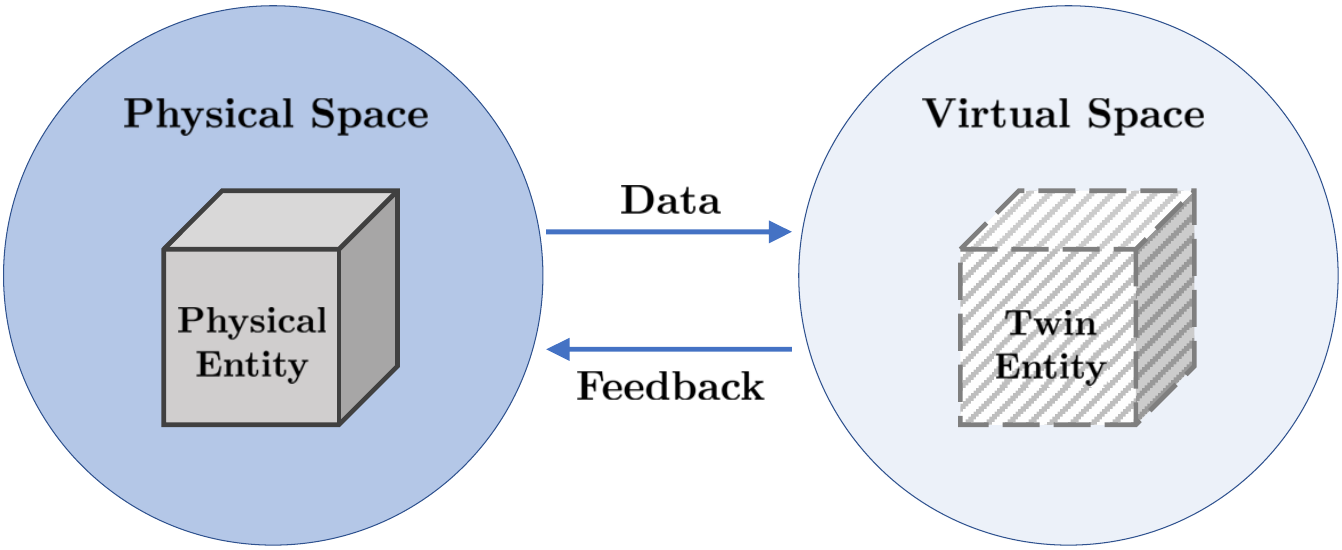}
		\label{fig.dt-architecture}}
	\hspace{4mm}
	\subfigure[MixedDT]
	{\includegraphics[scale=0.55]{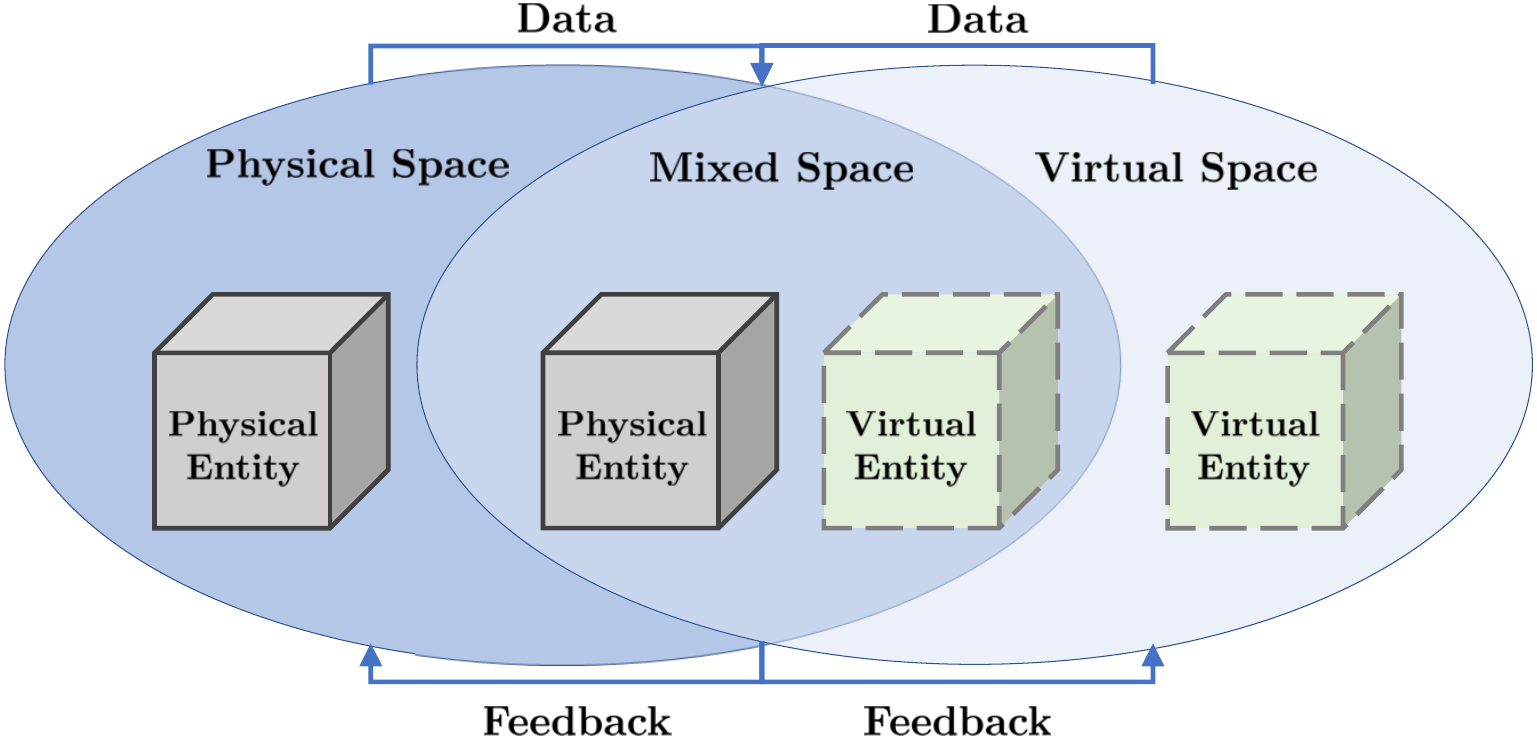}
		\label{fig.mdt-architecture}}
	\vspace{-1mm}
	\caption{Schematics for classical DT and mixedDT. (a) In classical DT, the virtual space is a digital replica of the physical space. (b) In mixedDT, the virtual space and the physical space is integrated into the mixed space, where physical and virtual entities coexist and interact with each other.
	}
	\label{fig.architecture}
\end{figure}

Similarly to the field tests, another significant problem of the existing miniature platforms lies in the limitations of the physical environment with respect to flexibility and scalability. It is non-trivial to extend the number or adjust the motion dynamics of experimental vehicles. To tackle this problem, researchers have tried to digitize the experimental platform, such as reconstructing the low-level vehicle dynamics~\cite{goldfain2019autorally} or even the entire physical platform~\cite{zayas2021digital} in the virtual space. Nevertheless, the physical space and the virtual space are essentially two separate experimental environments in~\cite{goldfain2019autorally,zayas2021digital}. To address this problem, Digital Twin (DT) is regarded as an effective tool, which is defined as a virtual representation of a physical asset enabled through data and simulators for real-time monitoring, prediction, optimization,  control, and decision making~\cite{grieves2014digital,rasheed2020digital,tao2018digital}. As shown in Fig.~\ref{fig.dt-architecture}, DT is composed of two main parts: physical entities in the physical space, and the counterparts of the physical entities (named as twin entities) in the virtual space~\cite{grieves2014digital}. 
Through synchronization and interaction of real-world activities with the virtual counterparts, DT promises to facilitate real-world experiments via virtuality~\cite{liao2021cooperative}. Very recently, several small-scale experiments have been carried out to verify the potential of DT in tests and demonstrations of CAV technologies, such as adaptive cruise control~\cite{wang2021digital2}, cooperative ramp merging~\cite{liao2021cooperative} and  intersection control~\cite{wang2021digital}. 
Nevertheless, due to DT's natural limitations in the virtuality-reality connections, where virtuality is commonly a digital reproduction of reality, the potential of the virtual space in developing CAV technologies has not been fully explored in the aforementioned work~\cite{wang2020digital,kumar2018novel,liao2021cooperative,wang2021digital}.

\subsection{Notion of Mixed Digital Twin (mixedDT)}

\begin{figure*}[t]
	\vspace{1mm}
	\centering
	\subfigure[]
	{\includegraphics[scale=0.41]{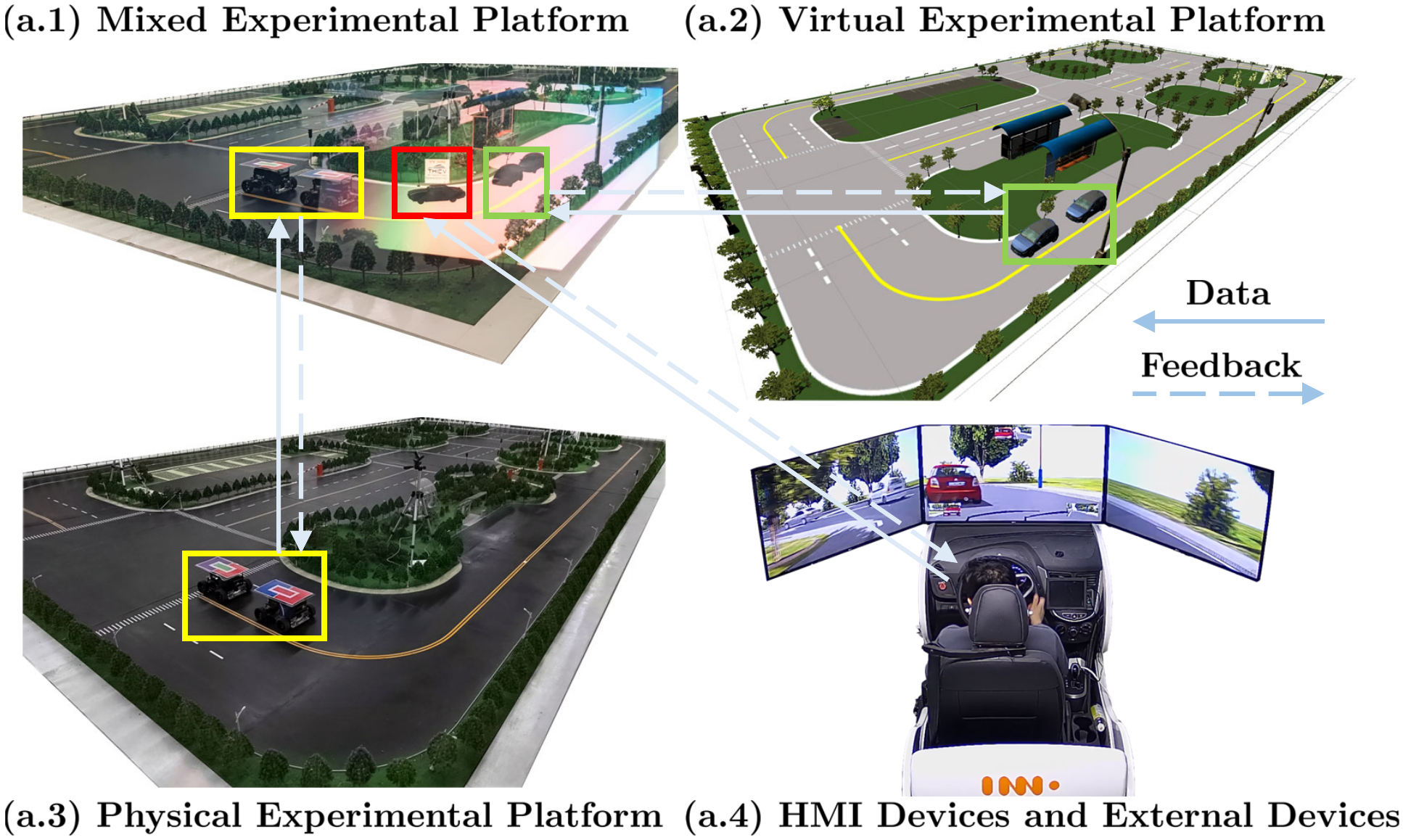}
		\label{fig.SICity-overview}}
	\hspace{4mm}
	\subfigure[]
	{\includegraphics[scale=0.5]{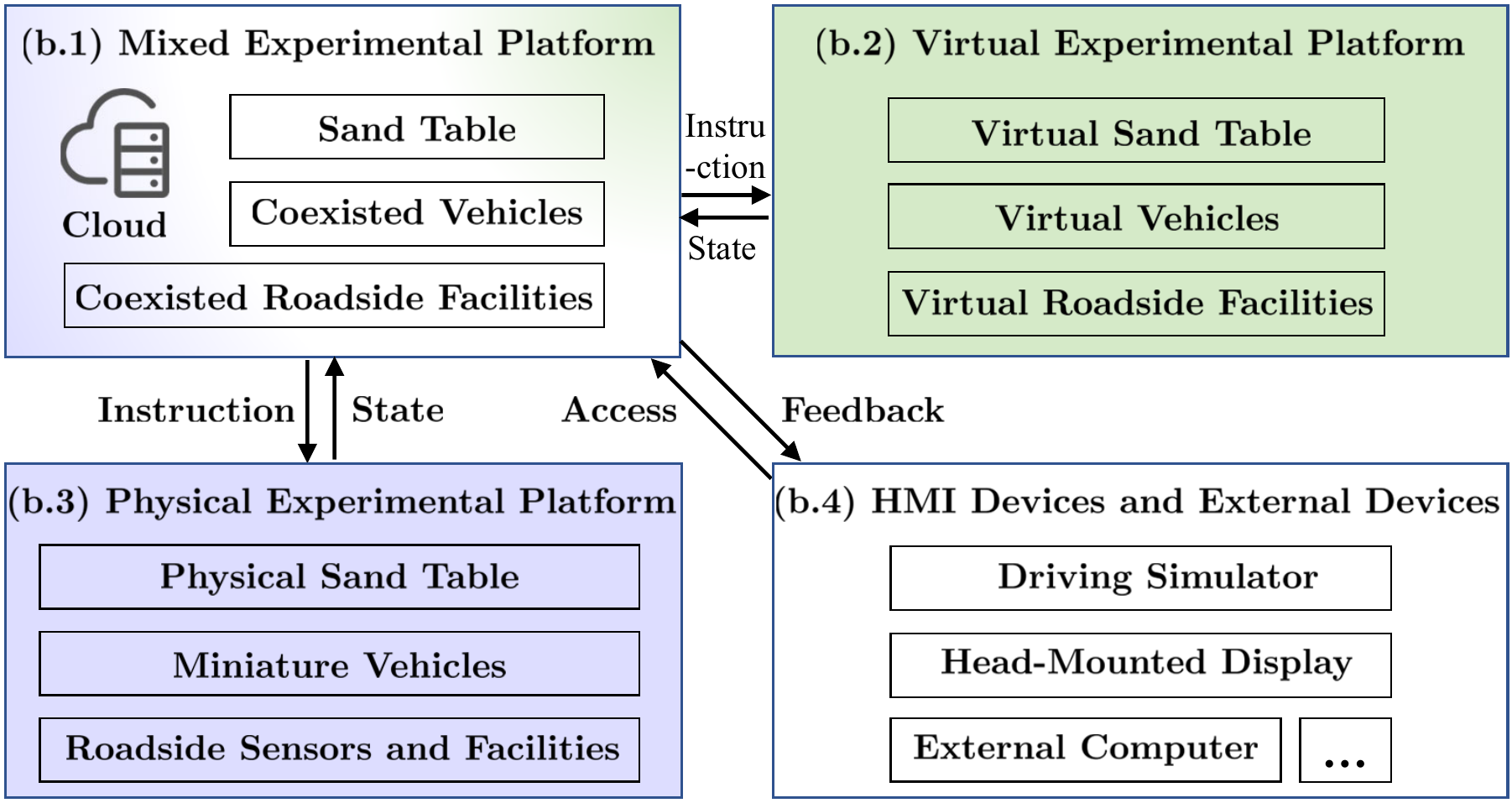}
		\label{fig.SICity-architecture}}
	\vspace{-1mm}
	\caption{Overview of MCCT. (a) Schematic of MCCT. In the mixed experimental platform, physical miniature vehicles, virtual vehicles and the HDV via the driving simulator coexist and interact with each other. The demonstration videos can be found in \url{https://github.com/dongjh20/MCCT}. (b) Diagram for the architecture of MCCT.
	}
	\label{fig.SICity}
\end{figure*}

In this paper, we extend the notion of DT and introduce a new concept of Mixed Digital Twin (mixedDT). As illustrated in Fig.~\ref{fig.mdt-architecture}, mixedDT consists of three main spaces: physical space, virtual space, and mixed space. 
The meanings of physical space and virtual space are consistent with those in the classical DT~\cite{glaessgen2012digital,grieves2014digital,grieves2017digital,tao2018digital}, while the newly added mixed space stands as an integration of physical space and virtual space, where physical and virtual entities coexist and interact in real-time. The mixed space can be observed via specialized display equipment, such as Microsoft HoloLens~\cite{tepper2017mixed}, and an illustration based on our MCCT is shown in Fig.~\ref{fig.SICity}(a.1).
Note that the virtual entity in mixedDT is not a digital replica or counterpart of the physical entity, \ie, twin entity in classical DT, but stands as its own in the virtual space without a corresponding physical counterpart. Therefore, virtual entities have the potential to interact with physical entities in the mixed space.

A related concept is Mixed Reality, which is defined as a type of simulators combining both virtual and physical objects to create a hybrid of the virtual and physical spaces~\cite{ohta2014mixed,flavian2019impact}. 
Indeed, Mixed Reality has also been recently adopted in CAV testing and evaluation~\cite{feng2020safety,wang2020augmented,szalai2020mixed}. 
MixedDT introduces the idea of Mixed Reality into classical DT, and bridges the physical space and the virtual space as an integrated whole by establishing the mixed space, where the physical entities and the virtual entities coexist and interact. Along this direction, the flexibility and scalability of DT in CAV validations can be greatly enhanced.
In practical implementation, physical entities can be twinned into the virtual space, and then aligned and aggregated with the data of virtual entities, thus indirectly realizing the coexistence and interaction of physical and virtual entities in the mixed space.

\subsection{Contributions}
In this paper, we introduce our miniature platform MCCT, which is established based on the notion of mixedDT and serves for experimental validations of vehicle-road-cloud integration. As shown in Fig.~\ref{fig.SICity-overview}, MCCT consists of four parts: physical experimental platform, virtual experimental platform, mixed experimental platform, and Human-Machine Interface (HMI) devices and external devices. The physical platform provides a scaled real-world test environment, while the virtual platform brings flexible design for the quantity, motion behavior and dynamics of vehicles. 
Particularly, MCCT allows for cross-platform experiments: the miniature vehicles in the physical experimental platform, the virtual vehicles in the virtual experimental platform and the HDV via the driving simulator can coexist and interact with each other in the mixed experimental platform. The real-time operation can be visualized projected to physical space and virtual space. 
Our contributions and the features of MCCT are specified as follows.

\begin{itemize}
	\item The notion of mixedDT is proposed via integrating Mixed Reality into DT, which facilitates CAV experiments with flexible extensions of various virtual entities,  and real-time interaction between physical and virtual entities. In classical DT~\cite{glaessgen2012digital,grieves2014digital,grieves2017digital,tao2018digital}, by contrast, virtual entities are digital replicas of physical entities. In particular, developed based on mixedDT, our platform MCCT overcomes the common limitations of existing miniature experimental platforms~\cite{la2012development,paull2017duckietown,karaman2017project}, such as vehicle quantity, motion behavior and low-level dynamics. 
	
	\item The three main platforms in MCCT (physical, virtual and mixed experimental platforms) allow for synchronous operation and cross-platform interaction based on mixedDT. Note that existing work mostly establishes physical and virtual platforms separately without deep integration~\cite{goldfain2019autorally,zayas2021digital}. In addition, the human driver factor can be introduced in a unified manner into all experimental environments, while existing work allows drivers to only interact with one particular environment~\cite{la2012development,wang2021digital}.
	
	\item Main components in vehicle-road-cloud integration are established in MCCT, including networked control, roadside localization, V2X  communication, smart roadside facilities, and cloud computing. In particular, a cloud unit is responsible for fusing data from various entities in different platforms, and assigning control instructions for them. 
	Cross-platform experiments are presented in this paper, where three types of vehicles (miniature vehicles, virtual vehicles, and HDVs) are organized into one platoon and coordinate with each other to dissipate traffic perturbations. More experiments in MCCT can be found for Cooperative Adaptive Cruise Control (CACC)~\cite{yang2021multi}, data-driven control in mixed traffic~\cite{wang2022experimental}, and multi-lane vehicle formation control~\cite{cai2021experimental}.
	
	\item Finally, MCCT has a high degree of flexibility and scalability with respect to further development through the modular platform design and the utilization of cross-language communication libraries.  Experiments can be carried out directly via any personal computer rather than platform-specific computers as before~\cite{stager2018scaled,hyldmar2019fleet,zayas2021digital}. Moreover, almost all major programming languages are supported, \eg, C/C++, Java, Python, MATLAB.
\end{itemize}

The rest of this paper is organized as follows. Section~\ref{sec.2} presents the overall architecture of the platform based on mixedDT. Section~\ref{sec.3} introduces the three main experimental platforms, and Section~\ref{sec.4} shows HMI and external devices. Experimental setup and results are presented in Section~\ref{sec.5}, and Section~\ref{sec.6} concludes this paper. Some preliminary demonstrations of our work have appeared in~\cite{yang2021multi}.

\section{MCCT Architecture based on MixedDT}
\label{sec.2}
In this section, we introduce the overall architecture of MCCT, which is a miniature implementation for the mixedDT concept. As shown in Fig.~\ref{fig.SICity-architecture}, MCCT consists of three main platforms: a physical experimental platform, a virtual experimental platform, and a mixed experimental platform, corresponding to the physical space, virtual space and mixed space in mixedDT, respectively. In addition, a fourth part of HMI devices and external devices is developed for visualization of and access to MCCT. 

The physical experimental platform, shown in  Fig.~\ref{fig.SICity}(a.3), represents the physical space in mixedDT. It is constructed on a sand table of size $9\,\mathrm{m} \times 5\,\mathrm{m}$, whose road has a $1:14$ ratio with respect to the real road environment. It contains multiple physical entities, including miniature vehicles running on the road, and roadside sensors and facilities. Miniature vehicles and roadside sensors, which are sensitive to communication latency, communicate directly with the cloud, while the communication between roadside facilities and the cloud is relayed through the unified control board. This communication setup is motivated by the idea of Multi-access Edge Computing (MEC)~\cite{hu2015mobile}, one of the core supporting technologies of the vehicle-road-cloud integration  system~\cite{xu2017internet,sasaki2018vehicle}. 
Precisely, the miniature vehicle communicates with the cloud through wireless connection. The miniature vehicle uploads its state data, such as velocity and acceleration, and receives instructions from the cloud. 
Four cameras installed on the ceiling serve as the roadside sensors and transmit the real-time captured images to the cloud through wired connection. 
Currently deployed roadside facilities include $73$ street lamps, $11$ traffic lights and $3$ lifting rods. These facilities  are electrically connected to a unified control board, which follows the control instructions from the cloud. 


The second main platform in MCCT is the virtual experimental platform, representing the virtual space in mixedDT. It is developed based on a $14:1$ digital replica of the physical sand table, thus having a consistent size as real-world roads; see Fig.~\ref{fig.SICity}(a.2) for demonstration. There exist multiple virtual entities on the virtual sand table, including virtual vehicles with simulation dynamics and virtual roadside facilities. 
The virtual experimental platform is developed based on a game engine, motivated by existing virtual platforms~\cite{wang2021digital2}. It communicates with the cloud through wired connection to ensure real-time performance. Precisely, the virtual vehicles and roadside facilities send their states to the cloud, which issues control instructions for them. 



The mixed experimental platform is inspired by the emerging Mixed Reality technology~\cite{flavian2019impact}, which overlaps the physical world and the virtual world into a holistic world in front of people’s eyes, and the physical objects and the virtual objects satisfy the real 3D projection relationship. It corresponds to the mixed space in mixedDT and is directly operating on the cloud. The mixed experimental platform acts as the most distinctive part of MCCT compared with existing miniature platforms; see Fig.~\ref{fig.SICity}(a.1) for illustration. The physical experimental platform, the virtual experimental platform, and the HMI devices and external devices in MCCT are all ``mixed'' into this mixed experimental platform, through which they indirectly interact with each other. Particularly, miniature vehicles and virtual vehicles from the physical and virtual experimental platforms respectively can coexist and interact with each other in a shared environment, \ie, the mixed space. In practical implementation, the cloud aligns and aggregates the data of miniature vehicles and virtual vehicles into a unified sand table environment in the virtual space. 
Miniature vehicles are twinned into the virtual space and interact with virtual vehicles indirectly via their digital counterparts, \ie, twin vehicles. This process is similar to DT in CAV applications~\cite{liao2021cooperative}.
In addition, physical and virtual roadside facilities could also coexist in the mixed experimental platform, enhancing the flexibility for road infrastructures.

Finally, we deploy HMI devices and external devices for visualization of and access to the mixed experimental platform. Specifically, the HMI devices include a Mixed Reality head-mounted display (Microsoft HoloLens) and a driving simulator. HoloLens receives necessary information and sends human gesture control commands to the cloud through wireless connection. In addition, HoloLens can also provide a visualization of the mixed experimental platform projected to physical space, as demonstrated in Fig.~\ref{fig.SICity}(a.1). Similarly, the driving simulator receives necessary information from the cloud, and sends the data from a human driver's driving behavior, such as steering wheel angle and accelerator pedal signal, to the cloud via wired connection. 
As for external devices, we allow personal computers and devices from other platforms to access MCCT via wired or wireless communication with the cloud. 

\begin{remark}[Combination of DT and Mixed Reality]
The classical DT notion has been recently adopted for autonomous driving experiments~\cite{liao2021cooperative} and driver-in-the-loop simulations~\cite{wang2021digital2}. In addition, preliminary attempts have also been made to use Mixed Reality~\cite{wang2020augmented,feng2020safety,szalai2020mixed}, for tests and evaluation of CAV technologies. In this paper, we combine the notion of DT and Mixed Reality, leading to an extended concept of mixedDT, and build an experimental platform for  experimental validations of vehicle-road-cloud integration.
\end{remark}

\section{Implementation Details of MCCT}
\label{sec.3}

In this section, we present the construction details of the three major platforms, and explain how mixedDT and vehicle-road-cloud integration is implemented in MCCT. 

\subsection{Physical Experimental Platform}

As shown in Fig.~\ref{fig.shapan}, the physical sand table is a scaled road transportation system with miniature vehicles and roadside sensors and facilities. In terms of road structure, there exist two-way four-lane  roads in the middle and two-way two-lane roads in both sides of the sand table. This setup supports the most common traffic scenarios in CAV research, such as platoon control~\cite{li2022cooperative}, on/off ramp control~\cite{liao2021cooperative}, intersection control~\cite{xu2018distributed} and traffic network management~\cite{liu2021reservation}. In the following, we present implementation details of miniature vehicles and roadside sensors and facilities.

\subsubsection{Miniature Vehicle}
\begin{figure}[t]
	\vspace{0mm}
	\centering
	\subfigure[]
	{\includegraphics[scale=0.45]{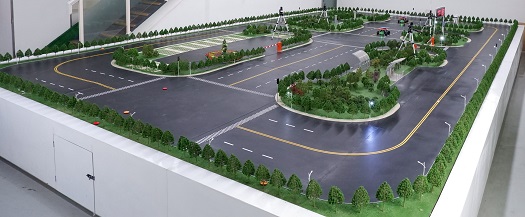}
		\label{fig.shapan}}
	\hspace{-1mm}
	\subfigure[]
	{\includegraphics[scale=0.70]{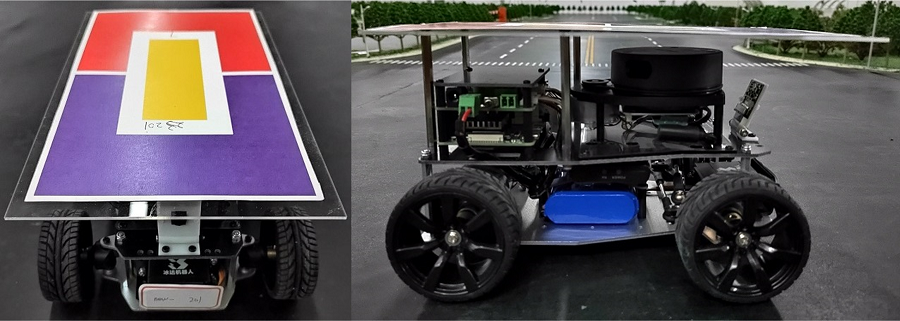}
		\label{fig.xiaoche}}
	\subfigure[]
	{\hspace{-10mm}\includegraphics[scale=0.53]{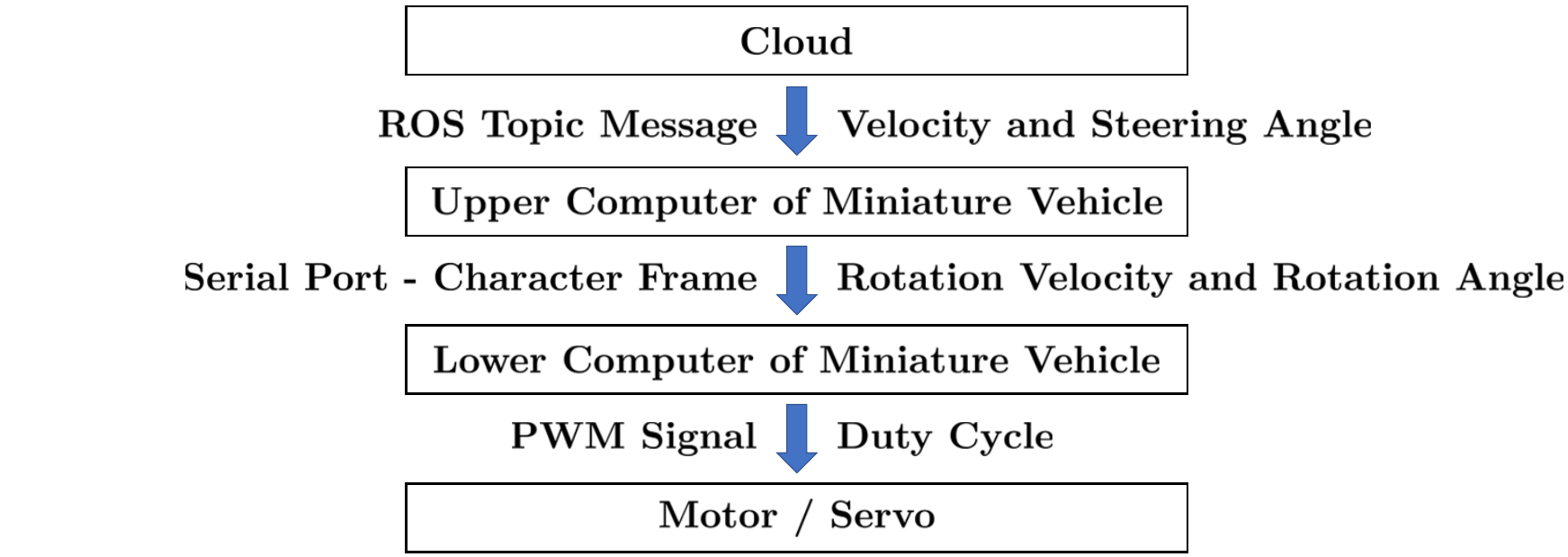} 
		\label{fig.control-architecture}}
	\caption{Physical experimental platform. (a) The physical sand table of size $9\,\mathrm{m} \times 5\,\mathrm{m}$. (b) The miniature vehicle, where a color block is pasted to the top. (c) Schematic of networked control structure of the miniature vehicle.
	}
	\label{fig.physical platform}
\end{figure}

As a major component in  vehicle-road-cloud integration systems~\cite{li2020Principles,chu2021cloud}, CAVs are represented by the motion and control of miniature vehicles in MCCT.
The miniature vehicle used on the sand table is shown in Fig.~\ref{fig.xiaoche}. There are currently nine identical vehicles. The vehicle is of size $215\,\mathrm{mm} \times 190\,\mathrm{mm} \times 125\,\mathrm{mm}$  
and of weight $1.4\,\mathrm{kg}$. Its wheelbase is $140\,\mathrm{mm}$, and the wheel diameter is $60\,\mathrm{mm}$. The maximum speed of stable operation is $1\,\mathrm{m/s}$, and the maximum front wheel steering angle is $40$\degree.

The upper computer of the vehicle is a Raspberry Pi 4 Model B with $1.5\,\mathrm{GHz}$ quad-core ARM Cortex A72 micro-processor and $2\,\mathrm{GB}$ memory, while the Microcontroller Unit (MCU) of lower computer is STM32F103RBT6. The vehicle is rear-wheel drive and front-wheel steering with an Ackerman steering structure. Two DC brushed motors on the rear axle generate driving force with a response frequency of $200\,\mathrm{Hz}$. The servo drives the front wheels to steer.
The on-board sensors in the miniature vehicle include a Light Detection and Ranging (LiDAR), an Inertial Measurement Unit (IMU), a camera and a Hall encoder. The velocity of the motor can be obtained through the Hall encoder difference and then the vehicle speed can be obtained. 
The energy source of each vehicle is a lithium battery with a capacity of $1800\,\mathrm{mAh}$, which allows for up to $3$-hour driving. 
The real-time current and voltage of the battery can be measured, supporting energy consumption related research. Note that we paste color blocks on the top of each vehicle,  which serve for roadside localization and will be detailed later in the introduction to roadside sensors. 

One emerging technology in vehicle-road-cloud integration is networked control of CAVs. Particularly, we consider the recently proposed cloud control architecture~\cite{xia2019cloud,chu2021cloud} for miniature vehicles in MCCT, and the design is shown in Fig.~\ref{fig.control-architecture}. The upper computer of the vehicle subscribes to the control instruction published in the Robot Operating System (ROS) topic message form by the cloud, and then converts it into the six Bytes character frame signal of the rotation 
velocity of the motor and the rotation angle of the servo. Finally,  the lower computer receives commands through serial port from the upper computer, and  controls the motor and servo by Pulse Width Modulation (PWM) signal to generate desired velocity and steering angle of the miniature vehicle.

For implementation of vehicle-cloud communication, we utilize ROS, an open source robot development software library and toolkit, to achieve cloud-based control of miniature vehicles in MCCT. Due to the convenience of the ROS framework, the calculation and the execution of the control instructions are fully decoupled, bringing great flexibility for external access to MCCT. 

\begin{figure}[t]
	\vspace{1mm}
	\centering
	\subfigure[]
	{\includegraphics[scale=0.4]{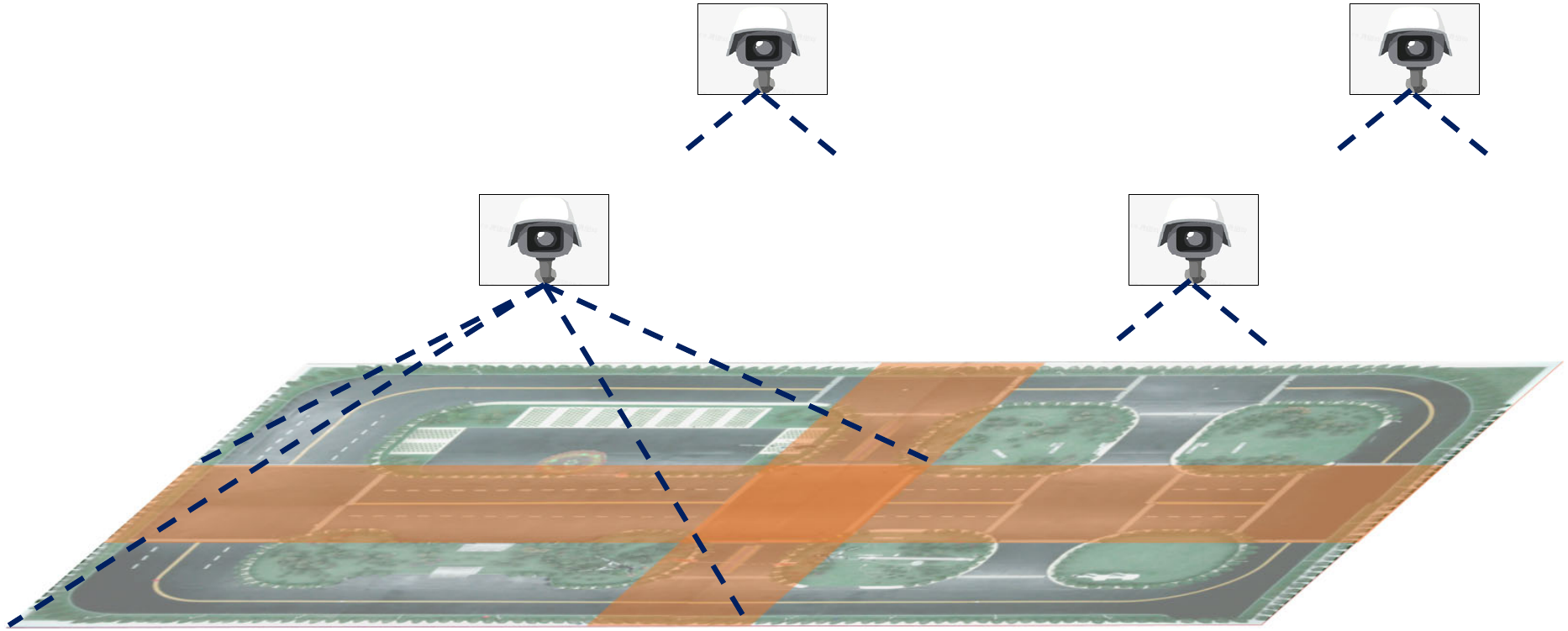}
		\label{fig.camera}}
	\hspace{1mm}
	\subfigure[]
	{\includegraphics[scale=0.4]{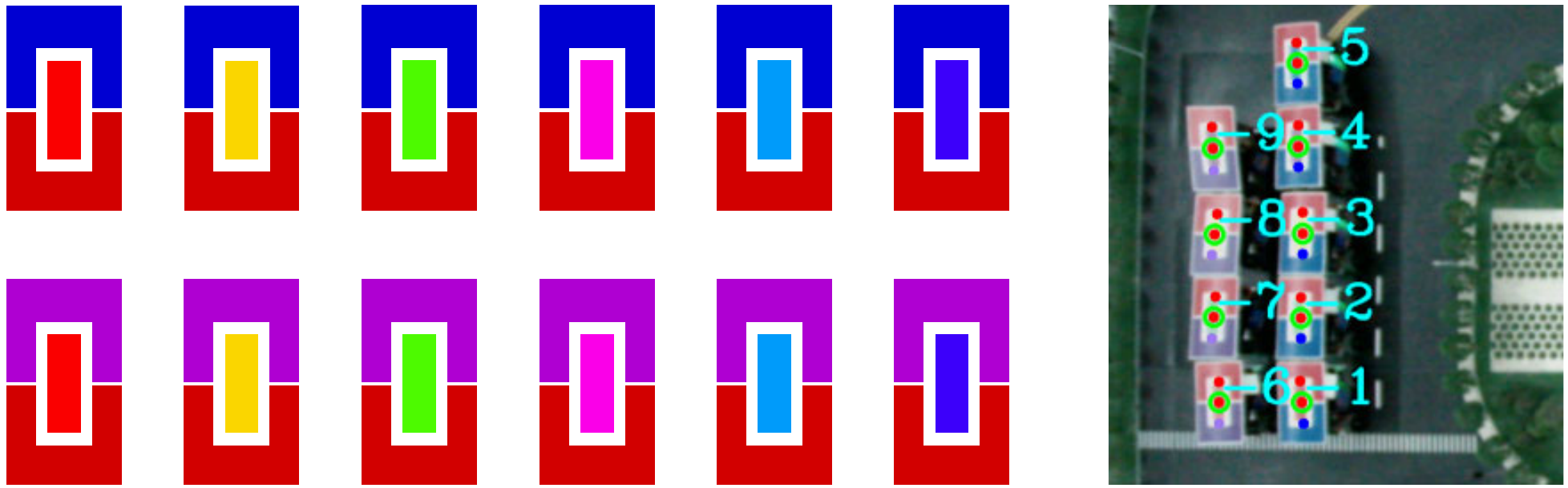}
		\label{fig.sekuai}}
	\vspace{-1mm}
	\caption{Roadside localization. (a) The $2 \times 2$ arrangement of the roadside cameras. (b) The design of color blocks, and an example of recognition results, where we use the numbers $1,2,\ldots,9$ to denote the vehicle identities.
	}
	\label{fig.virtual platform}
\end{figure}
\subsubsection {Roadside Sensors and Facilities} Roadside perception plays a critical role in vehicle-road-cloud integration, which removes the limitations of traditional on-board sensing technologies~\cite{chu2021cloud}. For representation of this component, we deploy four identical cameras as roadside sensors in MCCT,  as shown in Fig.~\ref{fig.camera}. The four cameras are laid out in a $2\times 2$  manner on the top of the entire physical platform, providing a top view of it. The camera's focal length is $204\,\mathrm{mm}$, operating frame rate is $20\,\mathrm{FPS}$, and the resolution is $1920\times1080$. 

The image captured from each camera is processed in real time for localization of each vehicle on the physical sand table by recognizing the color blocks on the top of each vehicle. The image processing related algorithms are developed based on the OpenCV library. 
The color block schemes and practical recognition results are shown in Fig.~\ref{fig.sekuai}. The color block consists of three parts by design. The rear part is fixed as red, and the front part is blue or purple, while for the middle part, different colors have been chosen. 
This design of color blocks allows us to recognize each vehicle's identity, and measure its position and heading via detecting the central point of each part of the color blocks; see the colored point in the right panel of Fig.~\ref{fig.sekuai}. In implementation, image transmission from the cameras to the cloud and the image processing in the cloud are executed in parallel, and each camera has been allocated with independent thread resources. 

The performance of this vision-based roadside localization is tested as follows. Based on data from $120$ uniformly sampled points on the sand table, the localization error is quantified, and the statistical results after fitting by Gaussian distribution are shown in Table~\ref{tab.positioning-error}.
We observe that large localization errors usually occur at the edge of the camera's field of view, but in general the localization error is relatively quite small compared to the size of the sand table. In addition, the image processing (computation) delay from receiving the image to output the localization result is also measured based on $6,000$ random tests. The results show that it obeys a Gaussian distribution with an average of $49.55\,\mathrm{ms}$ and a standard deviation of $1.39\,\mathrm{ms}$. 

\begin{table}[t]
\footnotesize
	\begin{center}
		\caption{Errors of Roadside Localization}
		\label{tab.positioning-error}
		\begin{threeparttable}
		\setlength{\tabcolsep}{4mm}{
			\begin{tabular}{ccc}
				\toprule
				& Gaussian Mean Value & Standard Deviation \\\hline
				X-direction & $15.18\,\mathrm{mm}$ & $19.65\,\mathrm{mm}$  \\
				Y-direction & $6.68\,\mathrm{mm}$ & $16.73\,\mathrm{mm}$  \\
				\bottomrule
		\end{tabular}}
		\begin{tablenotes}
		\footnotesize
		\item[1] The X and the Y directions represent the long side and the short side direction of the sand table, respectively.
		\end{tablenotes}
		\end{threeparttable}
	\end{center}
	\vspace{-4mm}
\end{table}

The deployment of the roadside facilities, including street lamps, traffic lights and lift rods, can support road-related research in vehicle-road-cloud integration such as smart street lamps and smart road gates. A unified control board is responsible for controlling all the roadside facilities, which receives  control instructions from the cloud through a pre-defined serial protocol, with a $9600$ baud rate.

\subsection {Virtual Experimental Platform}

\begin{figure}[t]
	\vspace{1mm}
	\centering
	\subfigure[]
	{\includegraphics[scale=0.82]{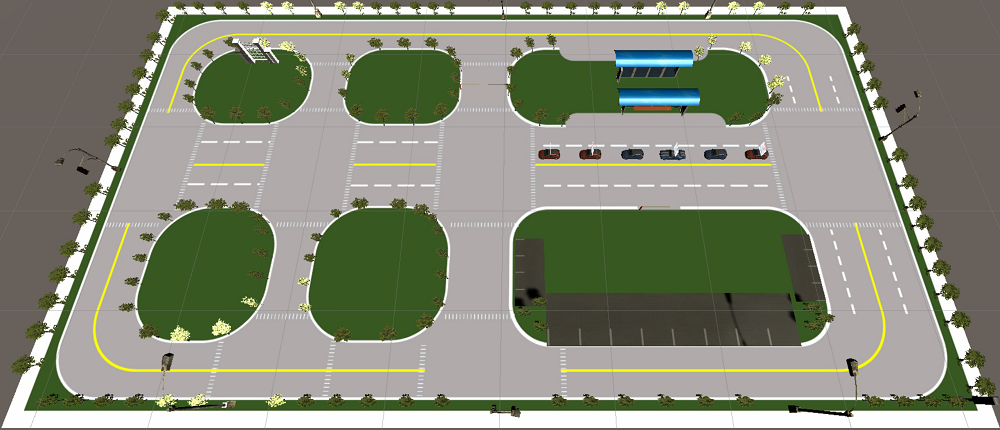}
		\label{fig.shuzishapan}}
	\subfigure[]
	{\includegraphics[scale=0.6]{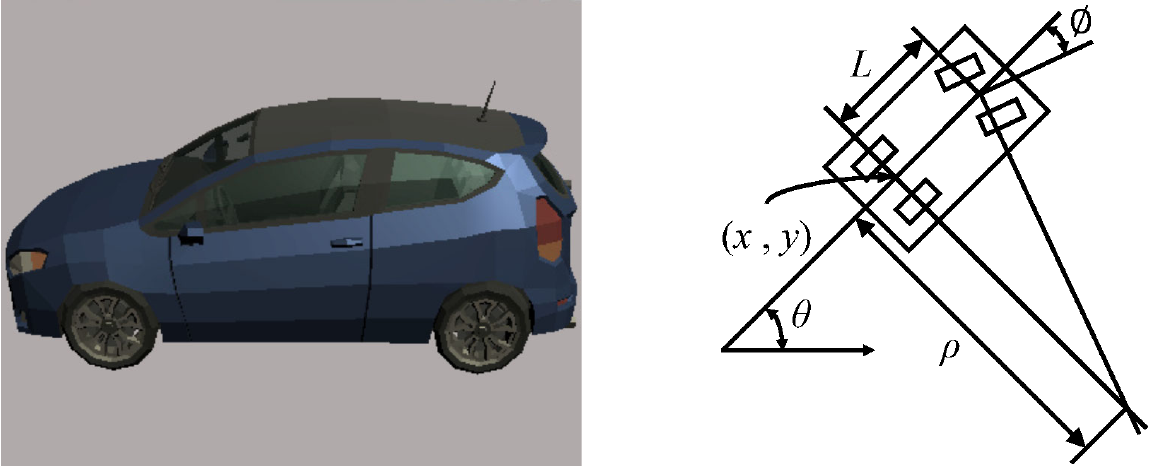}
		\label{fig.virtual vehicle}}
	\vspace{-1mm}
	\caption{Virtual experimental platform. (a) The virtual sand table. (b) The schematic and the dynamics model of the virtual vehicle.
	}
	\label{fig.yunche}
\end{figure}

Motivated by existing research on DT~\cite{wang2021digital2}, the virtual experimental platform runs in the Unity engine, and the Unity host is a workstation with Windows 10 system (Processor: Intel Core i7-10700K CPU @ 3.80 GHz x 16, Memory: $16\,\mathrm{GB}$). 
The physical sand table as shown in Fig.~\ref{fig.shapan} is completely reproduced in Unity with a $14:1$ scaling ratio; see  Fig.~\ref{fig.shuzishapan} for demonstration.

The virtual vehicles that only exist in the virtual experimental platform without physical counterparts are shown in the left panel of Fig.~\ref{fig.virtual vehicle}.
Classical DT commonly reproduces the physical vehicles in a digital platform (see, \eg,~\cite{goldfain2019autorally,zayas2021digital}), but due to the nature of the physical space, the vehicle quantity, motion behavior and dynamics are still fixed or limited. In mixedDT, by contrast, those limitations can be greatly relaxed, and particularly, the motion behavior and dynamics can be arbitrarily designed, and safety-critical scenarios can also be incorporated with low cost. This is typical in autonomous driving tests via Mixed Reality~\cite{ohta2014mixed,flavian2019impact,feng2020safety}, and in this paper we introduce this similar idea into our MCCT design.

\begin{figure}[t]
	\centering
	\subfigure[Step response for the longitudinal dynamics]
	{\includegraphics[scale=0.41]{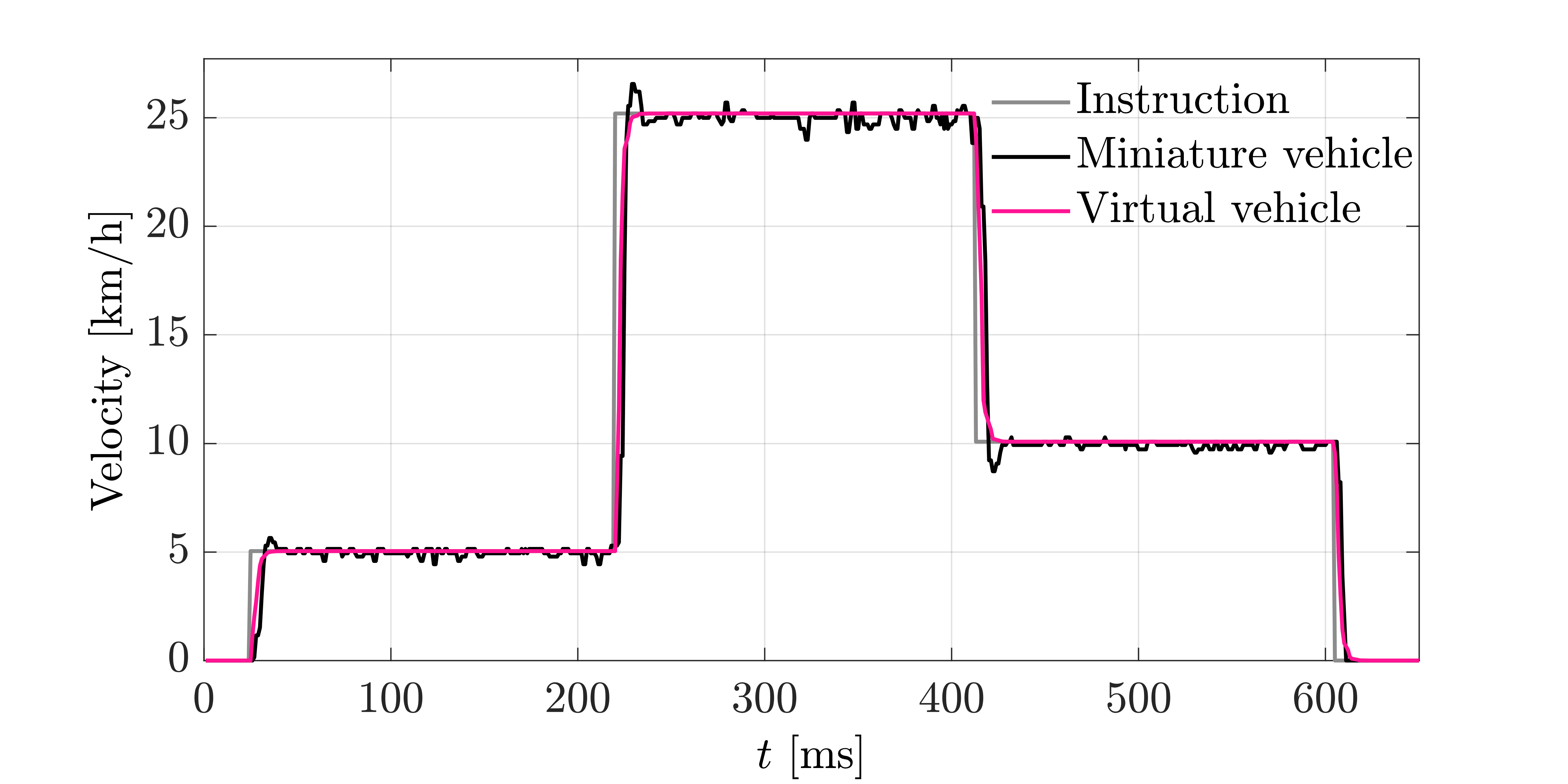}
		\label{fig.jieyue-sudu}}
	\subfigure[Step response for the lateral dynamics]
	{\includegraphics[scale=0.41]{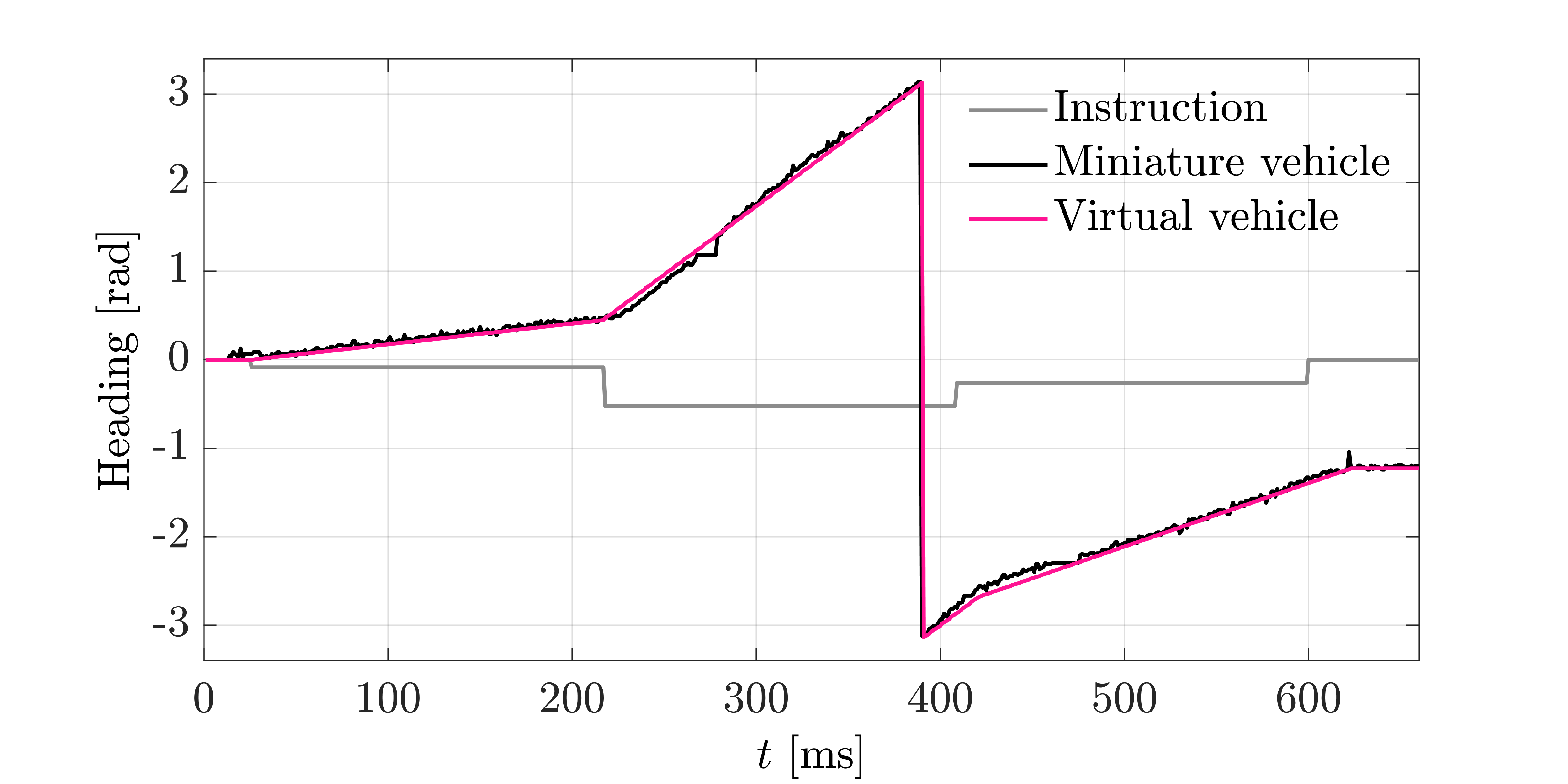}
		\label{fig.jieyue-zhuanjiao}}
	\vspace{-2mm}
	\caption{Longitudinal and lateral step response of the miniature and virtual vehicles. The velocity values have been unified to the virtual sand table environment, which has a consistent size as real-world roads. The mean deviation of the two response profiles is $0.34\,\mathrm{km/h}$ and $0.055\,\mathrm{rad}$ respectively, which is quite small. In (b), the instruction is front wheel steering angle.
	}
	\vspace{-3mm}
	\label{fig.jieyue}
\end{figure}

Since the dynamics of virtual vehicles  can be designed on demand, here we consider the typical bicycle model, given by
\begin{eqnarray}
	\begin{cases}
		\dot{x}(t)=v(t)\cdot \cos(\theta(t)),\\
		\dot{y}(t)=v(t)\cdot \sin(\theta(t)),\\
		\dot{\theta}(t)= \frac{v(t)}{L} \cdot \tan(\phi(t)),\\
		\dot{v}(t)=a(t),\\
	\end{cases}
	\label{equ.zixingche}
\end{eqnarray}
where $t$ denotes the current time, and the system parameter $L$ represents the wheelbase. In~\eqref{equ.zixingche}, the state variables $x$, $y$, $\theta$, $v$ represent the rear axle center's position, the yaw angle, and the speed of the vehicle, respectively, and the control inputs $a$, $\phi$ denote the acceleration and front wheel steering angle respectively; see the right panel in Fig.~\ref{fig.virtual vehicle} for demonstration. Without loss of generality, we design its dynamics behavior similar to the physical miniature vehicles. The longitudinal and lateral step responses of the miniature vehicle and the virtual vehicle are shown in Fig.~\ref{fig.jieyue}.  
It can be observed that both vehicles have quite similar dynamics behaviors, which confirms the potential of designing virtual vehicles with real-world dynamics behaviors. 

\begin{remark}[Virtual vehicles]
In classical DT, a virtual vehicle is a digital replica of a specific physical vehicle, which allows for vehicle control and optimization in the cloud. One distinct feature in mixedDT is that a virtual vehicle can stand on its own without a physical counterpart in the virtual space, and can further interact with physical vehicles in the mixed space. In this manner, the potential of virtual vehicles can be fully explored. Indeed, this notion of virtual vehicles is typical in CAV testing technologies via Mixed Reality~\cite{feng2020safety}. The interaction between virtual vehicles and physical vehicles will be elaborated in the following introduction to the mixed experimental platform of MCCT. 

\end{remark}

\subsection {Mixed Experimental Platform}
Motivated by the recent cloud/edge computing technologies~\cite{shi2016promise}, the cloud plays a critical role in the vehicle-road-cloud integration system, and is responsible for data fusion and system-wise control and optimization~\cite{li2020Principles,xia2019cloud,chu2021cloud}. Our mixed experimental platform, corresponding to the mixed space in mixedDT, is directly deployed on a cloud unit, which is the information transceiver center and control center, and provides a unified access for external devices to MCCT. Precisely, it is implemented in a cloud server with a processor of Intel Xeon Gold 5220R CPU @ 2.20GHz $\times$ 96 and a memory of 64 GB. 

Recall that the mixed space in mixedDT is an integration of the physical and virtual spaces, where physical and virtual entities coexist and interact in real-time. In MCCT, the cloud obtains the state information of miniature vehicles after fusing the measured data from roadside and on-board sensors, while the state information of virtual vehicles is naturally available to the cloud. After aligning and aggregating the data of both types of vehicles considering communication delay, the cloud projects the location and motion states of all the vehicles into a unified sand table in a shared environment, \ie, the mixed space. 
Based on the integrated information of the mixed space, the cloud makes necessary calculation and optimization, and sends the control instructions to the corresponding vehicles for execution. Consequently, the behaviors of miniature vehicles can truly affect those of virtual vehicles, and vice versa. Both types of vehicles from two different platforms coexist and interact with each other indirectly through the mixed experimental platform. Additionally, the roadside facilities in the physical and virtual platforms can also coexist and interact with each other in a similar approach. 

\begin{remark}[Insight of the mixed experimental platform] The mixed platform bridges the physical and virtual experimental platforms into a mixed (integrated) one, and realizes the coexistence and interaction of miniature vehicles and virtual vehicles. 
Consequently, the originally independent physical  and virtual experimental platforms achieve synchronous operation and cross-platform interaction. Due to the unified design of sand table environments in the physical and virtual platforms, the cloud can accurately and quickly align state data of the two types of vehicles and create the mixed space. In this manner, one can fully utilize the advantages of the two types of platforms and thus realize more potential applications than existing platforms (\eg,~\cite{stager2018scaled,goldfain2019autorally,zayas2021digital,wang2021digital2}). For visualization of the mixed experimental platform, we need to project it either into the virtual space via twinning miniature vehicles into the virtual space, or into the physical space via observing the virtual vehicles from a Mixed Reality display device; this will be detailed in Section~\ref{Sec:Visualization}. 
\end{remark}


\subsection{Communication and External Access}

\begin{figure}[t!]
	\centering
	\includegraphics[scale=0.68]{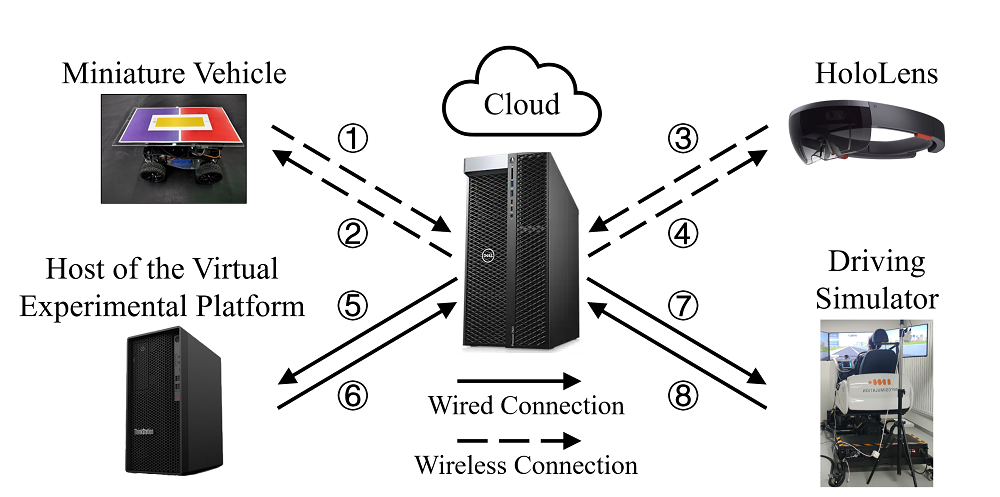}
	\vspace{-2mm}
	\caption{Schematic of the eight major communication links in MCCT. The miniature vehicle is in the physical experimental platform, while the HoloLens and the driving simulator are HMI devices. Wireless connection mode is $5\mathrm{GHz}$ WiFi, and wired connection mode is Ethernet.}
	\label{fig.shiyan-jiagou}
	\vspace{-2mm}
\end{figure}

\newcommand{\tabincell}[2]{\begin{tabular}{@{}#1@{}}#2\end{tabular}}  
\begin{table}[t!]
	\begin{center}
		\caption{Measurement of Communication Delays in the Eight Links (unit: $\mathrm{ms}$)}\label{tab.commuDelay}
		\begin{tabular}{ccccc}
			\toprule
			\tabincell{c}{Link}   &\tabincell{c}{Gaussian Mean} &\tabincell{c}{Standard Deviation}  &\tabincell{c}{ $99^\mathrm{th}$ Pencentile}  \\\hline
			\circled{1}/\circled{2}     & $1.33$ & $0.66$ & $2.86$ \\
			\circled{3}/\circled{4}     & $4.23$ & $1.72$ & $8.23$ \\
			\circled{5}/\circled{6}     & $0.38$ & $1.17$ & $3.09$ \\
			\circled{7}/\circled{8}     & $0.36$ & $2.74$ & $6.74$ \\
			\bottomrule
		\end{tabular}
	\end{center}
	\vspace{-4mm}
\end{table}

For communication between the cloud and other components in MCCT or external devices, we mainly utilize the open source cross-language communication library ZeroMQ, which supports almost all mainstream programming languages including C/C++/C\#, Java, Python, etc. For languages not supported by ZeroMQ, such as MATLAB, a lower-level socket communication library is employed. Hence, the cloud can communicate with various entities in MCCT. In addition, this approach also allows external devices driven by various programming languages to establish connections with the cloud and thus obtain access to MCCT. Particularly, experimenters can directly use their personal computers to interact with the cloud to conduct experiments rather than being limited to using the dedicated computer ``embedded" on the experimental platform. This design also demonstrates the flexibility and scalability of MCCT. 

A low value of communication delay plays a critical role in the normal operation of vehicle-road-cloud integration systems.  As shown in Fig.~\ref{fig.shiyan-jiagou}, there are mainly eight communication links in MCCT. Precisely, wireless connection mode is $5\mathrm{GHz}$ WiFi, and wired connection mode is Ethernet. The communication delay is measured after $7500$ random samples, and the statistical results after fitting by Gaussian distribution are shown in  Table~\ref{tab.commuDelay}.
As can be clearly observed, the communication delay in all the eight links in MCCT are quite low. Particularly, the communication delay between the miniature vehicles and the cloud obeys a Gaussian distribution with an average of $1.33\,\mathrm{ms}$ and a standard deviation of $0.66\,\mathrm{ms}$, which fully satisfies the requirements of cloud-based real-time control for vehicles~\cite{xia2019cloud,chu2021cloud,li2020Principles}.


\section{Visualization and Interaction for MCCT}
\label{sec.4}

In this section, we present the visualization methods of the mixed experimental platform, and elaborate the access to MCCT via HMI devices and external devices.

\subsection {Visualization of the Mixed Experimental Platform}
\label{Sec:Visualization}

One common problem in DT is the visualization of the twinned information from the cloud~\cite{wang2022mobility}. 
For the mixed experimental platform deployed on the cloud, which represents the mixed space of mixedDT, it is even more intractable to get a complete visualization, since it is essentially an integration of the physical and virtual spaces. In MCCT, we provide two approaches: 1) visualization projected to virtual space, and 2) visualization projected to physical space. 

\begin{figure}[t]
	\vspace{1mm}
	\centering
	\subfigure[]
	{\includegraphics[scale=0.54]{Figures/Unity-quanbu7.pdf}
		\label{fig.unity-keshihua}}
	\hspace{1mm}
	\subfigure[]
	{\includegraphics[scale=0.14]{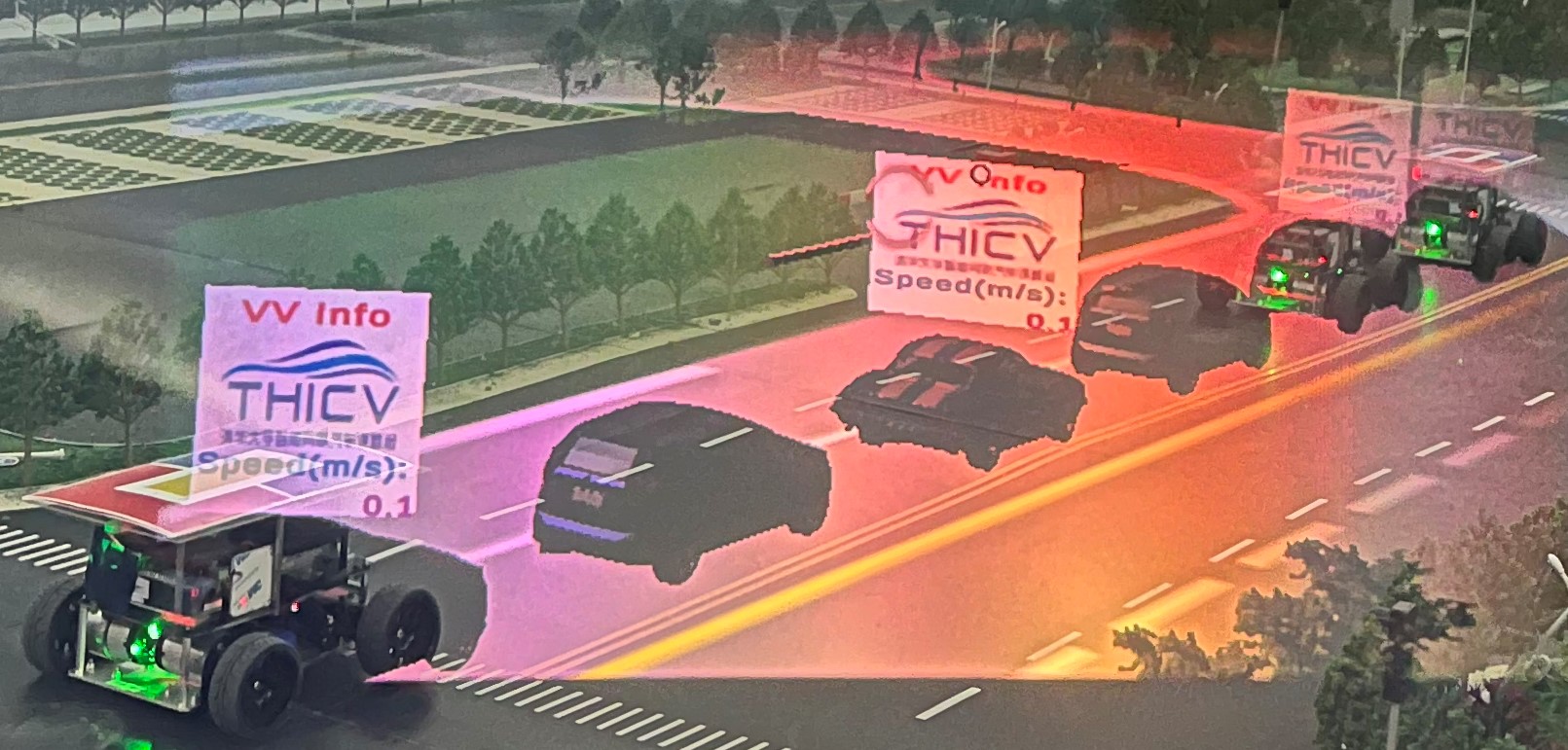}
		\label{fig.holo-keshihua}}
	\vspace{-1mm}
	\caption{Snapshots of the visualization results of the mixed experimental platform, where miniature vehicles, virtual vehicles, and an HDV coexist. (a) Visualization projected to virtual space. 
	(b) Visualization projected to physical space by HoloLens. The sequence of vehicles is the same as that in (a). 
	}
	\label{fig.keshihua}
\end{figure}

\subsubsection{Visualization Projected to Virtual Space}
Motivated by the visualization of DT in the CAV research field~\cite{wang2021digital2}, visualization projected to virtual space fully leverages the great flexibility of virtual vehicles. Recall that the quantity, motion behavior and dynamics of the virtual vehicles can be designed on demand. Accordingly, we create a corresponding number of virtual vehicles in the virtual experimental platform, and updates their states after necessary coordinate transformation based on the information of the miniature vehicles received in the cloud. These additionally created vehicles correspond to the twin vehicles in classical DT, which are digital replicas of the physical vehicles. A snapshot of this visualization method is shown in Fig.~\ref{fig.unity-keshihua}. The demonstrated vehicles consist of 1) virtual vehicles originally existing in the virtual experimental platform, 2) twin vehicles from the miniature vehicles in the physical experimental platform, and 3) a twin vehicle of the HDV via the driving simulator, which will be detailed in Section~\ref{Sec:HMI}.

\subsubsection{Visualization Projected to Physical Space} This visualization method utilizes the emerging Mixed Reality technology~\cite{flavian2019impact}. In MCCT, we employ the hardware device of HoloLens, which is one of the most prevailing head-mounted display devices for Mixed Reality; see Fig.~\ref{fig.shiyan-jiagou} for illustration. The Unity engine, which has been utilized for the development of the virtual experimental platform, also supports the software development of HoloLens. 
For implementation, HoloLens receives the necessary information, such as the states of the virtual vehicles, from the cloud via $5\mathrm{GHz}$ WiFi,  and then renders the virtual scene onto the lens, overlapping with the physical scene, which is directly visible to the naked eye. After manually aligning the virtual space with the physical space, one can see the real-time operation of the mixed experimental platform; see Fig.~\ref{fig.holo-keshihua} for the image obtained by directly photographing the lens of HoloLens in the physical space. Particularly, it can be seen that the miniature vehicles and the corresponding twin vehicles coincide at the same position, demonstrating the accuracy of coordinate alignment. 
Indeed, this ``mixed'' view is an intuitive illustration of the mixedDT concept via MCCT.

\subsection {HMI Design via HoloLens}
\label{Sec:HMI}


\begin{figure}[t!]
	\vspace{1mm}
	\centering
	\subfigure[]
	{\includegraphics[scale=0.38]{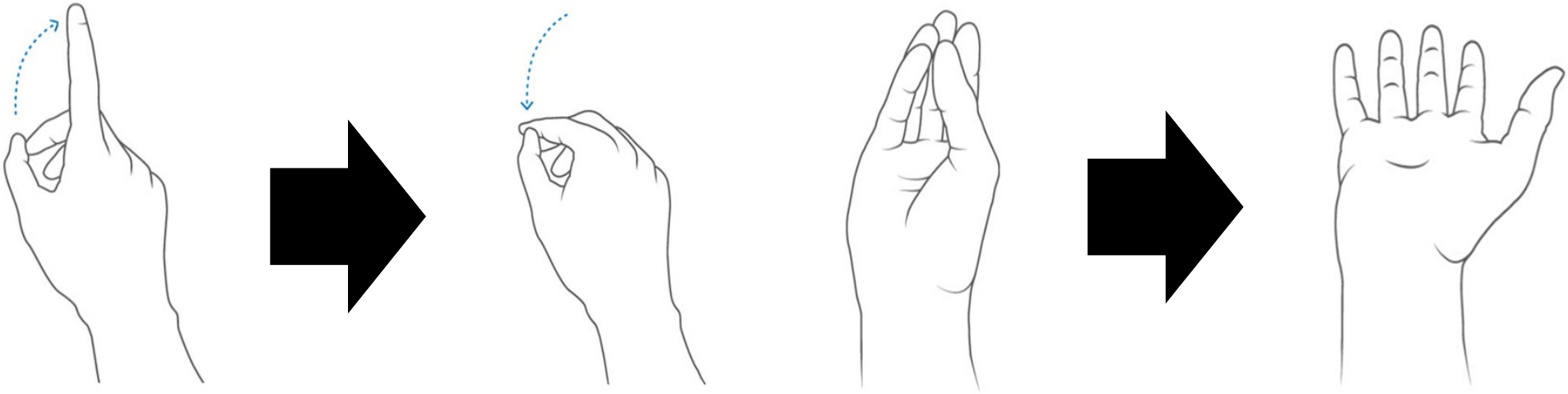}
		\label{fig.holo-shoushi}}

	\subfigure[]
	{\includegraphics[scale=0.8]{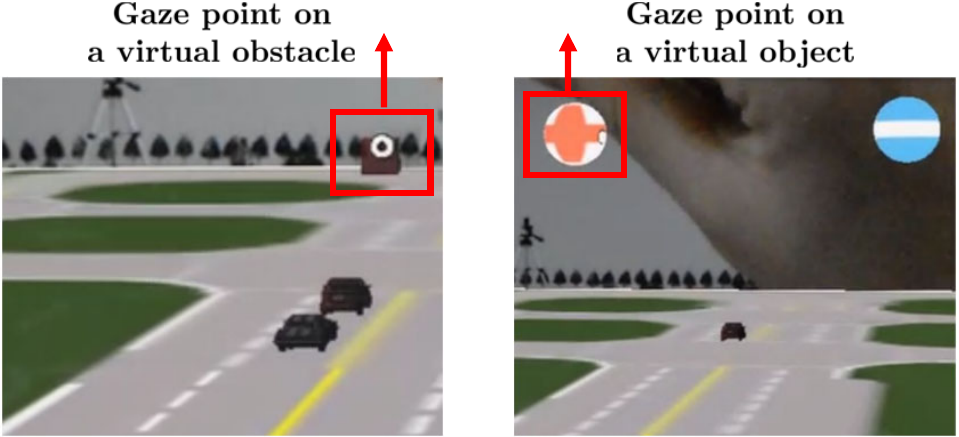}
		\label{fig.holo-app}}
	\vspace{-1mm}
	\caption{The recognizable gestures of HoloLens and implemented HMI applications. (a) The airtap gesture (left) and the boom gesture (right). (b) Using the airtap gesture to generate a virtual obstacle at specific location (left), and using the airtap gesture to control the velocity of a vehicle (right). 
	}
	\label{fig.shili-holo}
\end{figure}

HoloLens can not only support the  visualization of the mixed experimental platform, but also allow users to interact with MCCT in real time whilst enjoying an immersive 3D observation experience. It is equipped with multiple sensors and can accurately recognize human gestures; see, \eg, the airtap gesture  and the boom gesture in Fig.~\ref{fig.holo-shoushi}.  In practical implementation, the Mixed Reality ToolKit (MRTK) launched by Microsoft is used to convert user gestures to pre-defined control instructions. HoloLens uploads the instructions to the cloud via $5\mathrm{GHz}$ WiFi, which are then sent by the cloud to the corresponding entities for execution.

Here we present two interesting HMI applications in MCCT via HoloLens. As shown in the left panel in Fig.~\ref{fig.holo-app}, when a user shows an airtap gesture, HoloLens uploads the position of the user’s gaze point to the cloud as the location of an obstacle on the road after  coordinate transformation. In this way, a virtual obstacle is generated in MCCT, which only exists in the virtual space and does not exist in the physical space. This application allows users to put an obstacle in an arbitrary position at any time, which can be utilized for testing the obstacle avoidance algorithms. The other application is to change the velocity of one particular vehicle. When the user gazes at an interactive virtual object in the virtual space, an airtap gesture can trigger the predefined 
instruction, \eg, increasing the velocity of a vehicle; see the right panel in Fig.~\ref{fig.holo-app}. HoloLens uploads the velocity increase or decrease instruction to the cloud, and then the cloud sends it to the corresponding vehicle for execution after necessary conversion. This can be employed to impose external perturbations to one vehicle at any time during the experiments.

\subsection {HMI via Driving Simulator}

Human drivers still play a critical role in the upcoming vehicle-road-cloud integration systems, since there could still exist a high proportion of vehicles that are under human control in the near future~\cite{zheng2020smoothing,wang2021leading}. To capture this human driver factor into MCCT, we deploy a driving simulator of INNOSIMULATION. In practical implementation, we utilzie the supporting software of the driving simulator, \ie, SCANeR Studio, to measure the input signals from the human driver, including the steering wheel angle, the accelerator pedal signal and the brake pedal signal. The simulator communicates with the cloud via wired connection to ensure real-time performance (see Fig.~\ref{fig.shiyan-jiagou}), and exchanges the driver's control input signals and necessary data from the experimental platforms. In the cloud, the driver's control input signals are translated into standard vehicle control instructions, which are then sent to the corresponding vehicle. Particularly, we develop two specific driving modes in MCCT, including 1) driving a miniature vehicle on the physical sand table, 2) driving a virtual vehicle on the virtual sand table.

\begin{figure}[t]
	\vspace{1mm}
	\centering
	\subfigure[]
	{\includegraphics[scale=0.28]{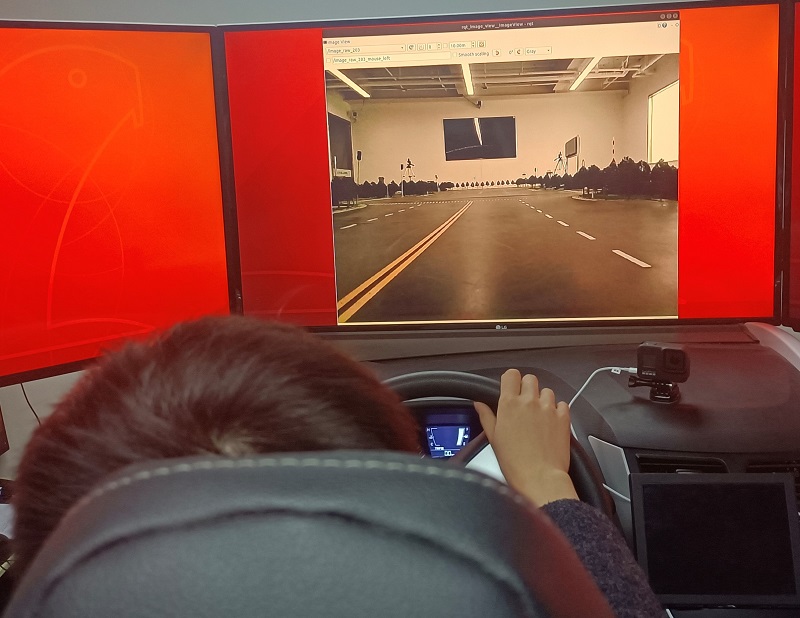}
		\label{fig.moniqi-kaizhenche}}

	\subfigure[]
	{\includegraphics[scale=0.185]{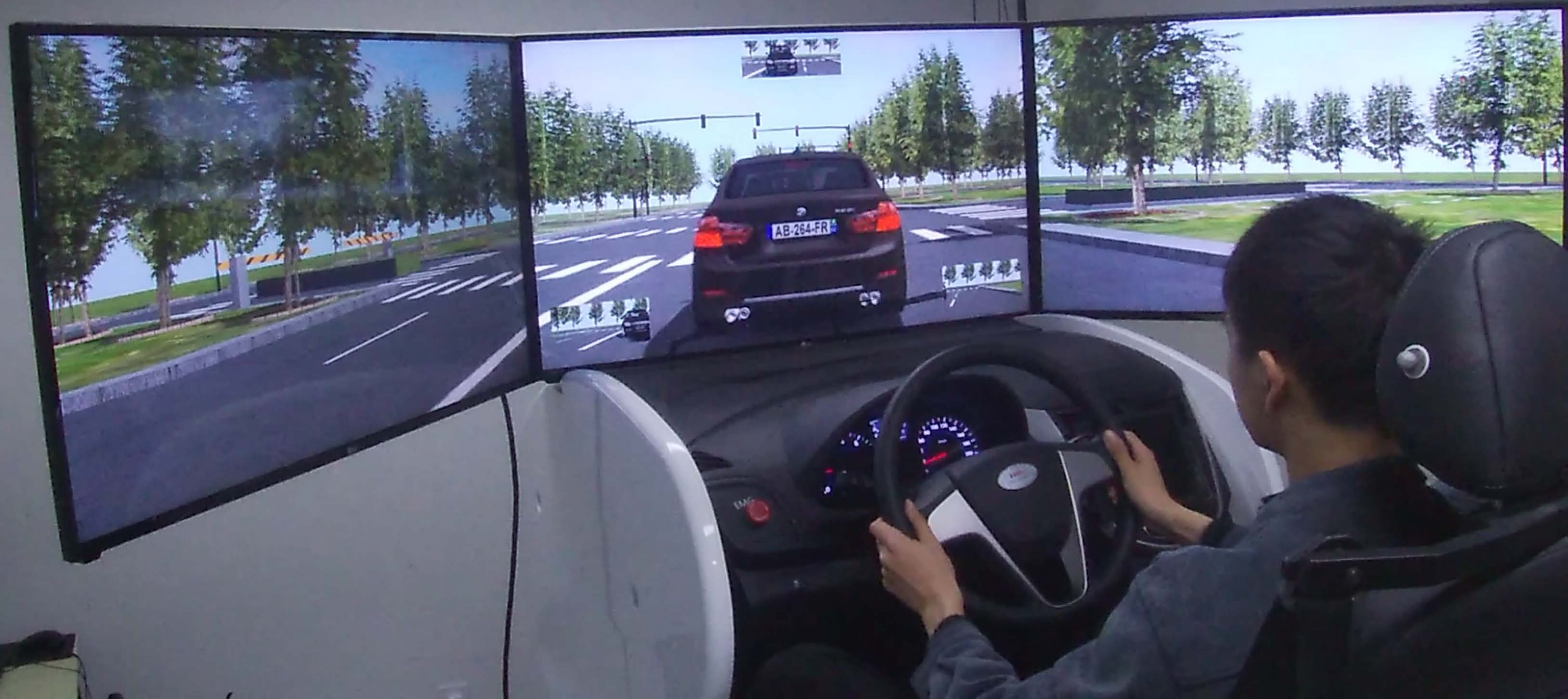}
		\label{fig.moniqi-kaiziche}}

	\vspace{-1mm}
	\caption{Snapshots of the human-in-the-loop experiments in two different modes: (a) Driving a miniature vehicle. (b) Driving a virtual vehicle.
	}
	\label{fig.moniqi-xianshi}
\end{figure}

\subsubsection{Driving a Miniature Vehicle}

Through the driving simulator, a human driver can drive a miniature vehicle based on the images captured by the on-board camera installed on the miniature vehicle; see Fig.~\ref{fig.moniqi-kaizhenche} for illustration. What the driver observes is the real-world scene in the physical sand table. This is typical in the existing miniature experimental  platforms based on a physical environment~\cite{la2012development}. However, since it is non-trivial to combine the virtual environment (particularly virtual vehicles) with the view from the on-board camera, this driving mode supports the research in the physical experimental platform only in MCCT.

\subsubsection{Driving a Virtual Vehicle}

The driving simulator also supports the control of a human driver on a virtual vehicle. Thanks to the convenience of Unity, the driver’s field-of-view of a virtual vehicle in the virtual space can be obtained by simply adding a camera object at an appropriate position on the vehicle object. Then, this view can be directly displayed to the human driver via screens in the driving simulator; see Fig.~\ref{fig.moniqi-kaiziche} for visualization. Similarly, this approach also provides access for the driver to control a twin vehicle of a miniature vehicle in the mixed experimental platform, whose motion is indeed captured in the physical sand table. Accordingly, this driving mode allows for the research on all the three experimental platforms in MCCT. 

\begin{remark}[Comparison of driving simulators with existing work]
Existing work mostly utilizes a driving simulator to control either a physical vehicle in the physical space~\cite{la2012development} or a virtual vehicle in the virtual space~\cite{wang2021digital,wang2020augmented}.
There lacks a unified method to incorporate human drivers into cross-platform experiments, \eg, transferring the human control of a virtual vehicle directly to a physical vehicle and vice versa. 
In MCCT, the potential of the virtual vehicles and the driving simulators are fully exploited thanks to mixedDT. The driver’s field-of-view of either a miniature vehicle or a virtual vehicle can be obtained, and the driver's control input can be executed on both types of vehicles. 
\end{remark}

\section{Experimental Validations}
\label{sec.5}
In this section, we present two platooning experiments to validate the effectiveness of MCCT. Particularly, we aim to demonstrate its capability for synchronous operation and cross-platform interaction by organizing different multi-source vehicles into a platoon in the mixed experimental platform.

\subsection{Experimental Setup}
\begin{figure}[t]
	\vspace{1mm}
	\centering
	\subfigure[]
	{\includegraphics[scale=0.42]{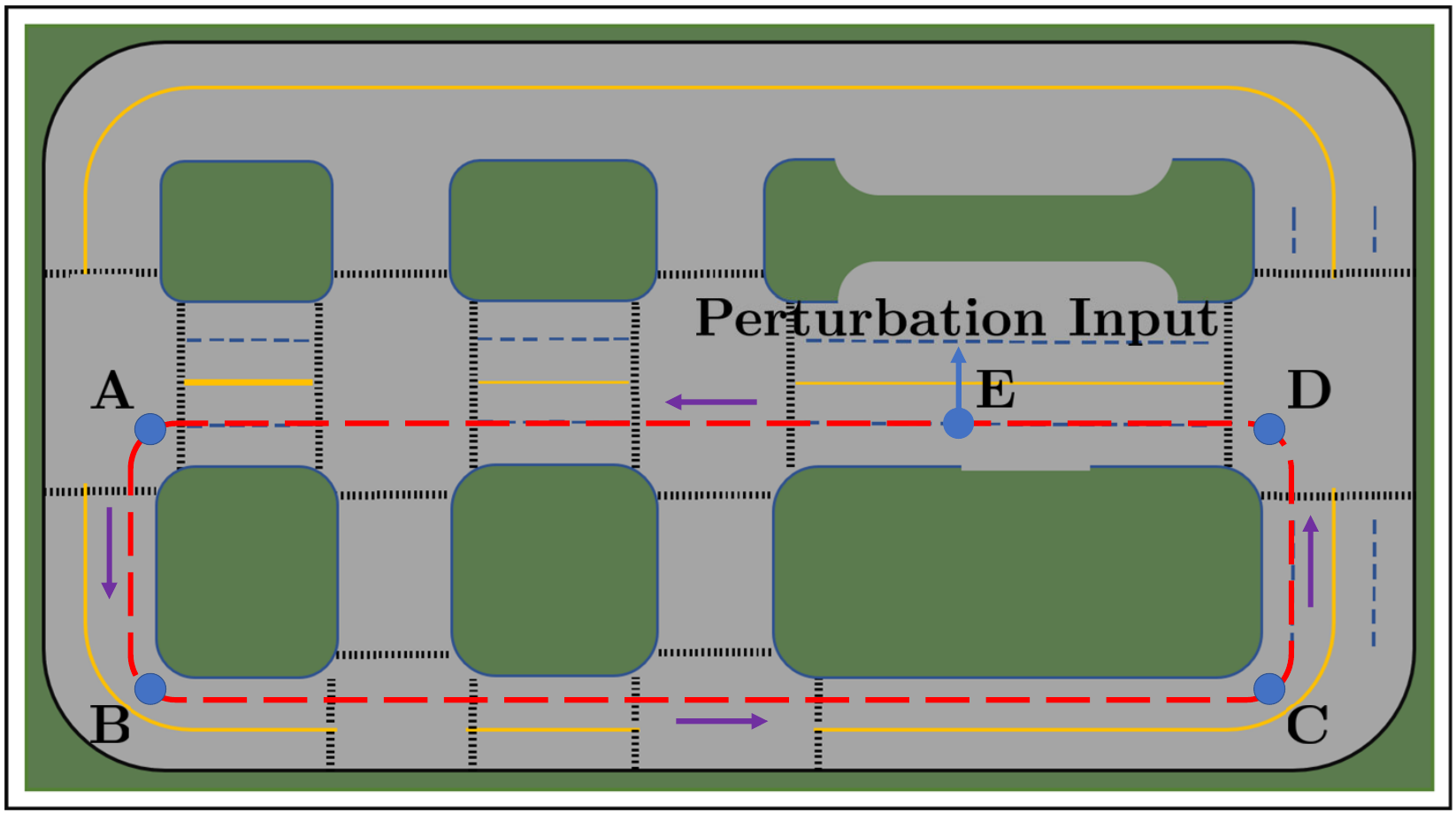}
		\label{fig.guiji}}
		
	\subfigure[]
	{\includegraphics[width=8.5cm]{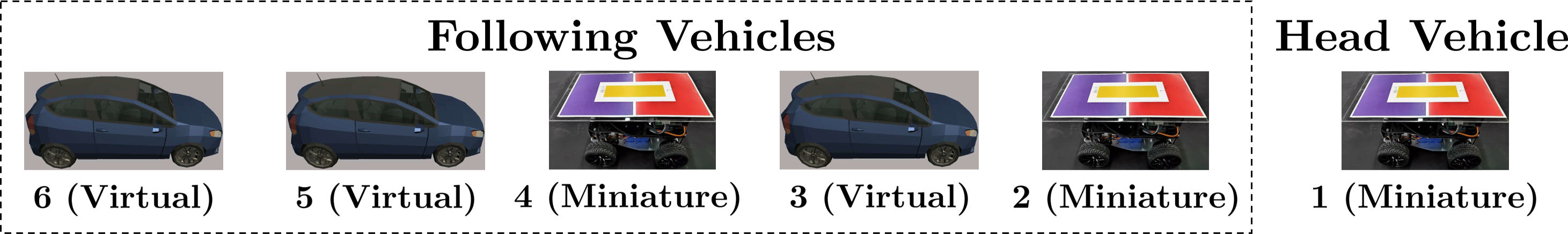}
		\label{fig.formation-hunhe}}
	
	\subfigure[]
	{\includegraphics[width=8.5cm]{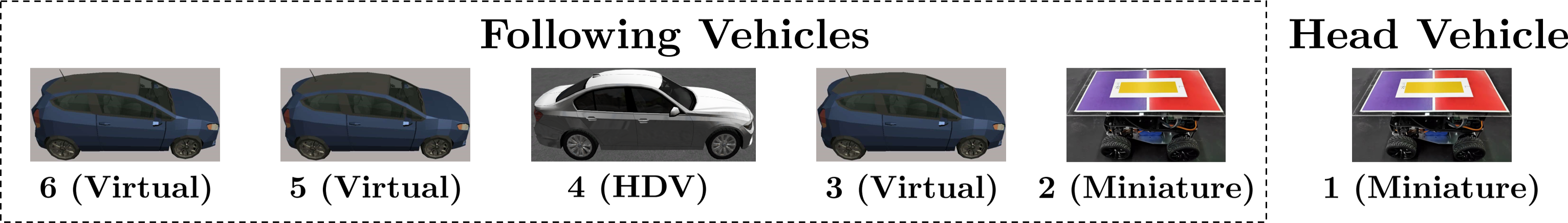}
		\label{fig.formation-quanhun}}
	\vspace{-1mm}
	\caption{Driving track and formations of the platoon. (a) The vehicles drive along the direction $A \rightarrow B\rightarrow C\rightarrow D\rightarrow A$. When reaching the $E$ point, the velocity of the head vehicle suffers from a sinusoidal disturbance. (b) The formation of the platoon for Experiment A. (c) The formation of the platoon for Experiment B. Particularly, the HDV is generated from the vehicle motion in the native supporting software of the driving simulator, \ie, SCANeR Studio. 
	}
	\label{fig.formation}
\end{figure}

A fleet of $6$ vehicles, consisting of one head vehicle and $5$ following vehicles, are deployed to run along the track $A \rightarrow B\rightarrow C\rightarrow D\rightarrow A$ in Fig.~\ref{fig.guiji} with a lap of about $17.5\,\mathrm{m}$. Among the $6$ vehicles, different multi-source vehicles are under consideration, including miniature vehicles, virtual vehicles and an HDV. The difference between the two experiments (Experiment A and Experiment B) lies in whether the HDV exists; see Fig.~\ref{fig.formation-hunhe} and Fig.~\ref{fig.formation-quanhun} for the specific formation. Precisely, in Experiment A, all the vehicles represent CAVs with a prescribed controller (including miniature vehicles and virtual vehicles), while in Experiment B, there is one vehicle under human control via the driving simulator.

The velocity profile of the head miniature vehicle (vehicle $1$ in Fig.~\ref{fig.formation-hunhe} and Fig.~\ref{fig.formation-quanhun}) is pre-designed. It maintains a fixed velocity of $0.3\,\mathrm{m/s}$, and a sinusoid perturbation occurs when it reaches the $E$ point shown in  Fig.~\ref{fig.guiji}, whose period is $3.5\,\mathrm{s}$ and amplitude is $0.1\,\mathrm{m/s}$. 
As for the control strategy of miniature vehicles and virtual vehicles, we consider a typical preview trajectory tracking controller~\cite{amer2017modelling} for lateral control. For longitudinal control, the following CACC controller is adopted~\cite{milanes2013cooperative}
\begin{equation}
	\label{equ.cacc}
	\begin{aligned}
		a_i(t)=&k_{p}\left((p_i(t)-p_{i-1}(t))-d_\mathrm{des}\right)+k_{v1}(v_{1}(t)-v_i(t)) \\ &+k_{v2}(v_{i-1}(t)-v_i(t)),
\end{aligned}
\end{equation}
where the control signal $a_i$ represents the desired acceleration of vehicle $i$ and $k_{p}$, $k_{v1}$ and $k_{v2}$ are all feedback gains. The state variables $p_{i}$ and $v_{i}$ denote the longitudinal position and velocity of vehicle $i$ respectively, and the system parameter $d_\mathrm{des}$ represents the desired inter-vehicle distance; see Table~\ref{tab.FeedbackGainSetup} for the parameter values. 
Recall that the executable longitudinal control instruction is the velocity as shown in Fig.~\ref{fig.control-architecture}, the desired acceleration $a_i(t)$ of each vehicle is converted to the velocity command signal $v_{i,\mathrm{cmd}}(t)$ in the cloud by
\begin{equation} \label{Eq:CommandVelocity}
    v_{i,\mathrm{cmd}}(t) = v_{i}(t-1) + a_i(t) \Delta t,
\end{equation}
where $v_{i}(t-1)$ represents the received  velocity of vehicle $i$ at the last time step $t-1$, and $\Delta t$ represents the time interval between $t$ and $t-1$. 

\begin{table}[t!]
	\begin{center}
		\caption{Parameter Setup in the CACC Controller~\eqref{equ.cacc}}\label{tab.FeedbackGainSetup}
		\begin{tabular}{ccccc}
			\toprule
			& $k_{p}$ & $k_{v1}$ & $k_{v2}$ & $d_\mathrm{des}\, [\mathrm{m}] $ \\\hline
			Miniature Vehicle & $0.25$ & $0.60$ & $0.60$  & $0.60$\\
			Virtual Vehicle & $0.10$ & $0.50$ & $0.50$  & $0.60$ \\
			\bottomrule
		\end{tabular}
	\end{center}
\end{table}


\begin{figure*}[t]
	\centering
	\subfigure[Experiment A]
	{\includegraphics[scale=0.48 ]{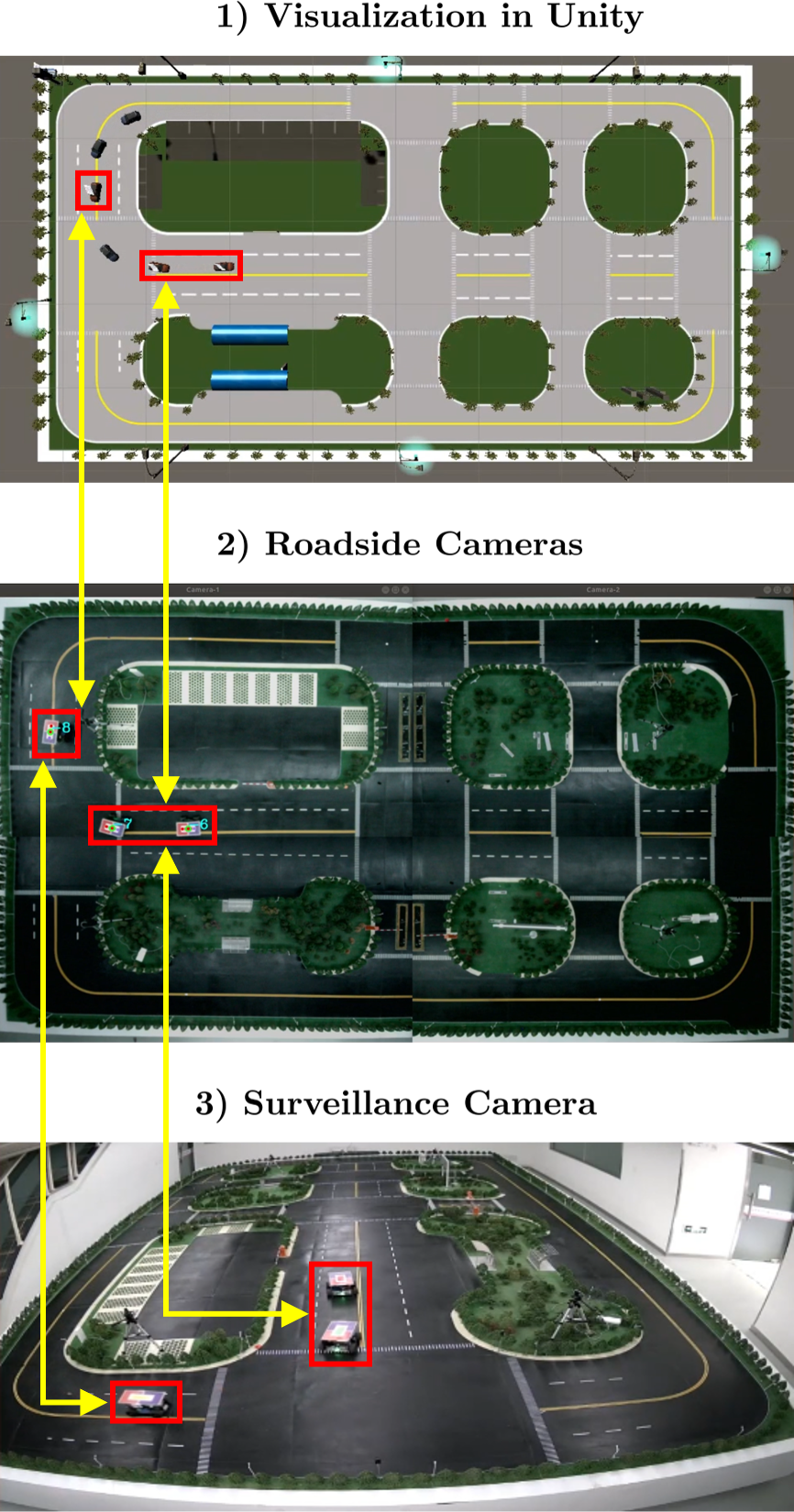}
	\label{fig.snapshot-hunhe}
	}
	\hspace{10mm}
	\subfigure[Experiment B]
	{\includegraphics[scale=0.48 ]{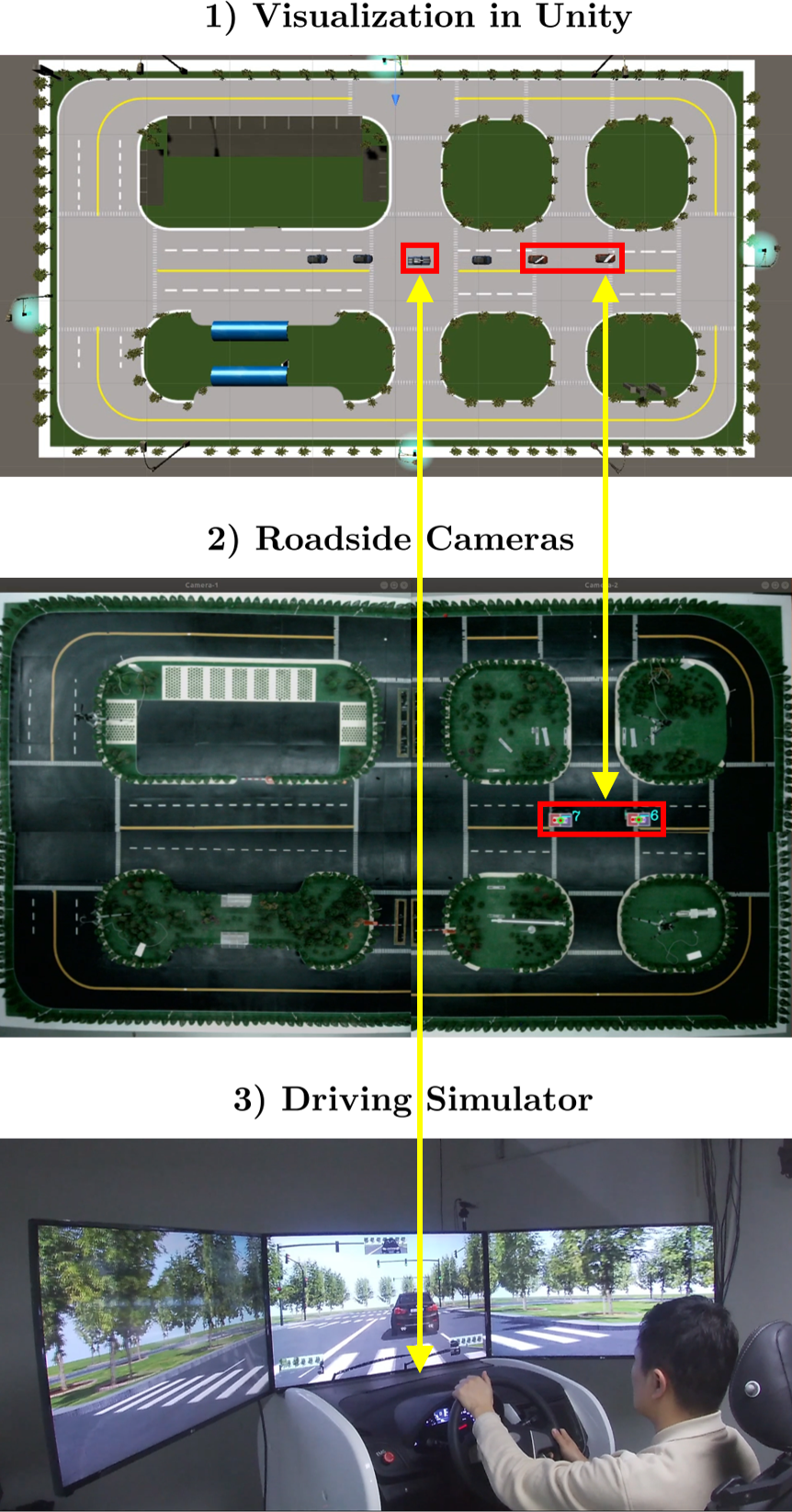}
	\label{fig.snapshot-quanhun}
	}
	\vspace{-1mm}
	\caption{Snapshots of experiments. The corresponding formations of the platoon are shown in Fig.~\ref{fig.formation-hunhe} and Fig.~\ref{fig.formation-quanhun} for Experiment A and Experiment B, respectively. Different multi-source vehicles interact with each other in the mixed space, and the visualization projected to virtual space in Unity is presented on the top panel. (a) As shown in the roadside cameras or the surveillance camera, there are three miniature vehicles running on the physical sand table. The other three vehicles are virtual vehicles.  (b) As shown in the roadside cameras, there are two miniature vehicles running on the physical sand table. One HDV is from the supporting software of the driving simulator. The other three vehicles are virtual vehicles. }
	\label{fig.snapshot}
\end{figure*}


\begin{figure*}[t]
	\centering
	\subfigure[Experiment A]
	{\includegraphics[scale=0.4]{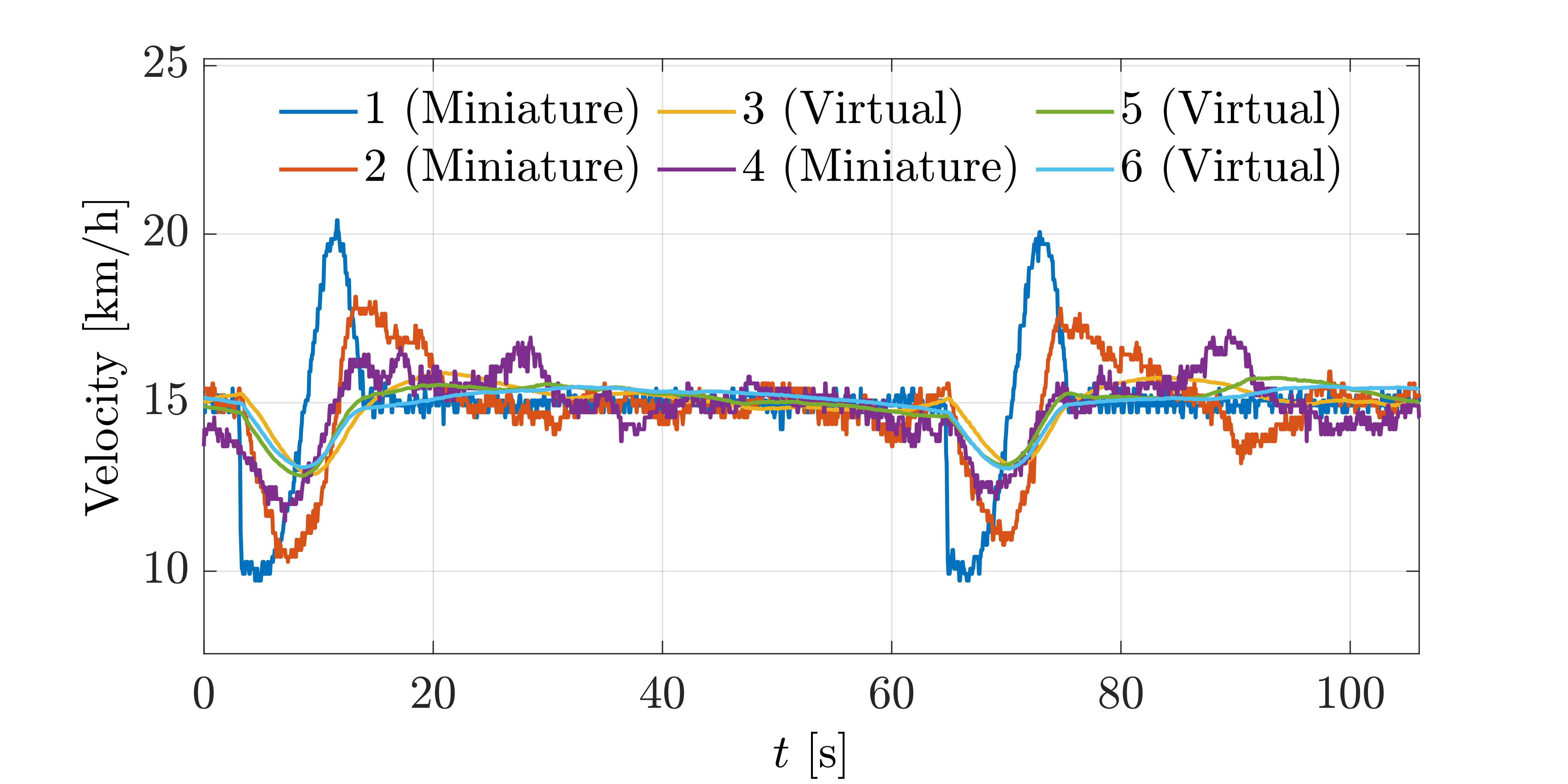}
		\label{fig.mixed-speedBig}}
	\subfigure[Experiment B]
	{\includegraphics[scale=0.4]{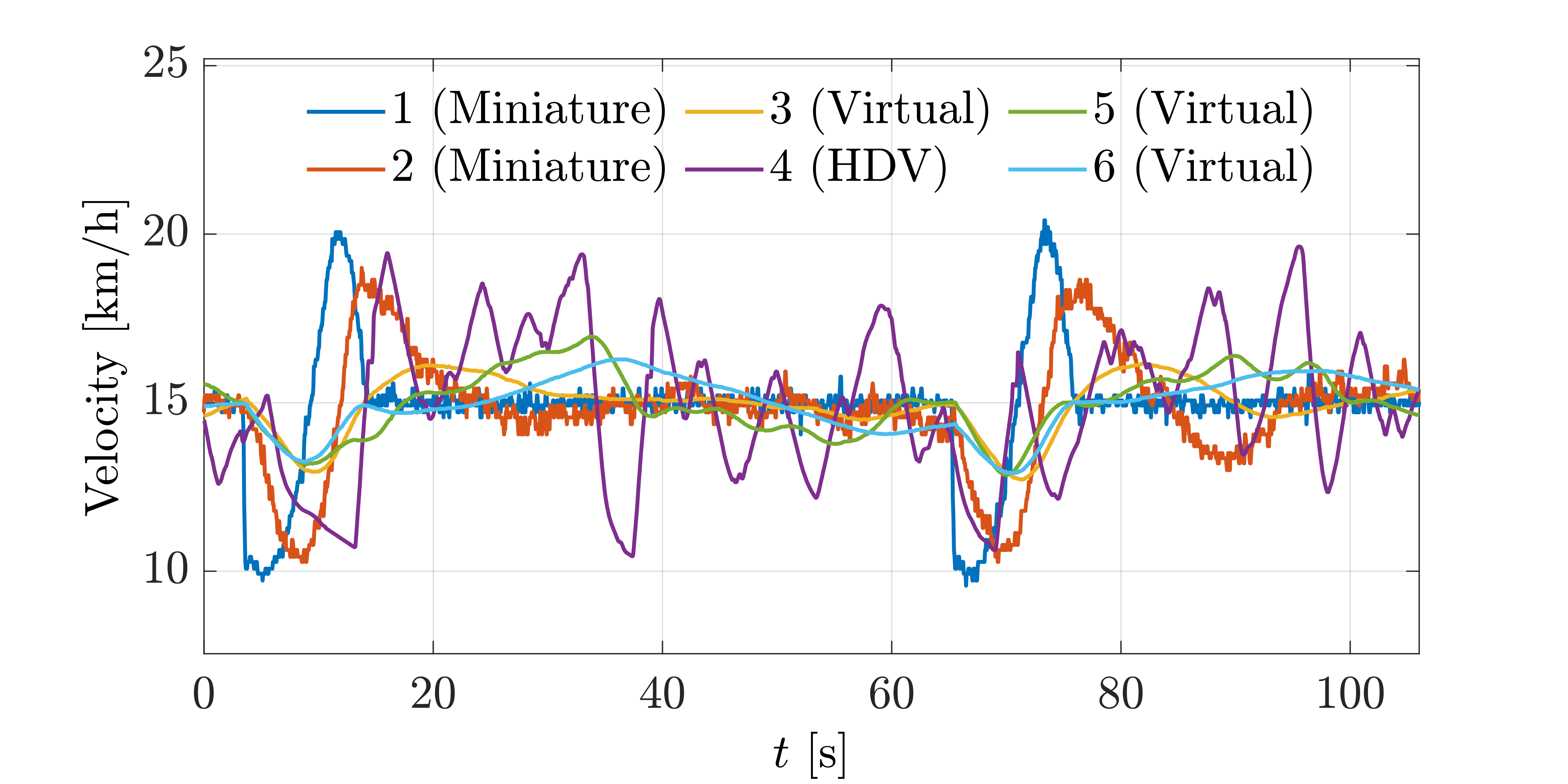}
		\label{fig.mixed-speed}}
	\vspace{-1mm}
	\caption{Speed profiles of all the vehicles in the experiments. Note that there is an HDV (magenta profile) in Experiment B, which has a relatively large oscillation. The velocity values have been unified to the virtual sand table environment, which has a consistent size as real-world roads.
	}
	\label{fig.speed}
\end{figure*}

\subsection{Experimental Results}

The snapshots of the two experiments are shown in Fig.~\ref{fig.snapshot}, and the recorded videos can be found in  \url{https://github.com/dongjh20/MCCT}.
The velocity profiles of all the vehicles are shown in Fig.~\ref{fig.speed}. Note that the velocity values are unified to the virtual sand table environment, which has a consistent size as real-world roads. 

It can be clearly observed that the platoon, composed of vehicles from multiple sources, run normally in the mixed experimental platform. In Experiment A, the CACC controllers  enable the CAVs to dampen the velocity fluctuation of the head vehicle, and achieve platoon stability.	In Experiment~B, since the fourth vehicle is controlled by a human driver via the driving simulator, its velocity has a larger oscillation amplitude, which captures the behavior of a particular driver into our experimental scenario. However, the fluctuations of the two following 
virtual vehicles are still well mitigated by the CACC controllers. We also observe that the velocity  profile of the virtual vehicle is smoother than that of the physical vehicle. This is due to the unknown noise and disturbance in the real-world physical environment, which cannot be completely replicated in the digital simulators. These experimental results validate the effectiveness of synchronous operation and cross-platform interaction of the three experimental platforms and HMI devices and external devices, which have been deeply integrated in MCCT.


\section{Conclusion}
\label{sec.6}
In this paper, we introduce our experimental platform MCCT, which is established based on the notion of mixedDT and serves for experimental validations of vehicle-road-cloud integration. The four parts of MCCT (physical, virtual, and mixed experimental platforms, and HMI devices and external devices) achieve synchronous operation and cross-platform interaction. Particularly, we have validated this capability of MCCT by organizing different vehicles from multiple sources into a platoon for cooperative driving in the mixed experimental platform. 
This demonstrates the great flexibility and scalability of MCCT, providing more opportunities for validating multi-vehicle cooperation and vehicle-road-cloud integration technologies. 

One interesting future direction is to add more HMI devices such as driving simulators to allow more human drivers involved in the CAV validations. Considering that the roadside control unit also plays a significant role in vehicle-road-integration systems, another interesting topic is to utilize intelligent intersection signals and ramp metering lights for verifying vehicle-road cooperative control technologies. More algorithm verification work using MCCT is also in progress, such as multi-vehicle coordination at unsignalized intersections, and eco-approaching at signalized interactions.



\ifCLASSOPTIONcaptionsoff
  \newpage
\fi



%

\bibliographystyle{IEEEtran}
\bibliography{IEEEabrv,mybibfile}

\begin{thebibliography}{10}
\providecommand{\url}[1]{#1}
\csname url@samestyle\endcsname
\providecommand{\newblock}{\relax}
\providecommand{\bibinfo}[2]{#2}
\providecommand{\BIBentrySTDinterwordspacing}{\spaceskip=0pt\relax}
\providecommand{\BIBentryALTinterwordstretchfactor}{4}
\providecommand{\BIBentryALTinterwordspacing}{\spaceskip=\fontdimen2\font plus
\BIBentryALTinterwordstretchfactor\fontdimen3\font minus
  \fontdimen4\font\relax}
\providecommand{\BIBforeignlanguage}[2]{{%
\expandafter\ifx\csname l@#1\endcsname\relax
\typeout{** WARNING: IEEEtran.bst: No hyphenation pattern has been}%
\typeout{** loaded for the language `#1'. Using the pattern for}%
\typeout{** the default language instead.}%
\else
\language=\csname l@#1\endcsname
\fi
#2}}
\providecommand{\BIBdecl}{\relax}
\BIBdecl

\bibitem{xia2015cloud}
Y.~Xia, ``Cloud control systems,'' \emph{IEEE/CAA Journal of Automatica
  Sinica}, vol.~2, no.~2, pp. 134--142, 2015.

\bibitem{li2020Principles}
K.~Li, J.~Li, X.~Chang, B.~Gao, Q.~Xu, and S.~Li, ``Principles and typical
  applications of cloud control system for intelligent and connected
  vehicles,'' \emph{Journal of Automotive Safety and Energy}, vol.~11, no.~3,
  p. 261, 2020.

\bibitem{chu2021cloud}
W.~Chu, Q.~Wuniri, X.~Du, Q.~Xiong, T.~Huang, and K.~Li, ``Cloud control system
  architectures, technologies and applications on intelligent and connected
  vehicles: a review,'' \emph{Chinese Journal of Mechanical Engineering},
  vol.~34, no.~1, pp. 1--23, 2021.

\bibitem{li2022cooperative}
K.~Li, J.~Wang, and Y.~Zheng, ``Cooperative formation of autonomous vehicles in
  mixed traffic flow: Beyond platooning,'' \emph{IEEE Transactions on
  Intelligent Transportation Systems}, vol.~23, no.~9, pp. 15\,951--15\,966,
  2022.

\bibitem{xu2018distributed}
B.~Xu, S.~E. Li, Y.~Bian, S.~Li, X.~J. Ban, J.~Wang, and K.~Li, ``Distributed
  conflict-free cooperation for multiple connected vehicles at unsignalized
  intersections,'' \emph{Transportation Research Part C: Emerging
  Technologies}, vol.~93, pp. 322--334, 2018.

\bibitem{liu2021reservation}
S.~Liu, Y.~Feng, and G.~Wu, ``Reservation-based network traffic management
  strategy for connected and automated vehicles: A multiagent system
  approach,'' in \emph{2021 IEEE International Intelligent Transportation
  Systems Conference (ITSC)}.\hskip 1em plus 0.5em minus 0.4em\relax IEEE,
  2021, pp. 2150--2155.

\bibitem{guanetti2018control}
J.~Guanetti, Y.~Kim, and F.~Borrelli, ``Control of connected and automated
  vehicles: State of the art and future challenges,'' \emph{Annual reviews in
  control}, vol.~45, pp. 18--40, 2018.

\bibitem{xia2019cloud}
Y.~Xia, C.~Yan, X.~Wang, and X.~Song, ``Intelligent transportation
  cyber-physical cloud control systems,'' \emph{ACTA Automatica Sinica},
  vol.~45, no.~1, pp. 132--142, 2019.

\bibitem{jia2015survey}
D.~Jia, K.~Lu, J.~Wang, X.~Zhang, and X.~Shen, ``A survey on platoon-based
  vehicular cyber-physical systems,'' \emph{IEEE communications surveys \&
  tutorials}, vol.~18, no.~1, pp. 263--284, 2015.

\bibitem{feng2021intelligent}
S.~Feng, X.~Yan, H.~Sun, Y.~Feng, and H.~X. Liu, ``Intelligent driving
  intelligence test for autonomous vehicles with naturalistic and adversarial
  environment,'' \emph{Nature communications}, vol.~12, no.~1, pp. 1--14, 2021.

\bibitem{Mcity-website}
Mcity, ``Mcity official website,'' Website, 2022, https://mcity.umich.edu/.

\bibitem{xu2021system}
S.~Xu, R.~Zidek, Z.~Cao, P.~Lu, X.~Wang, B.~Li, and H.~Peng, ``System and
  experiments of model-driven motion planning and control for autonomous
  vehicles,'' \emph{IEEE Transactions on Systems, Man, and Cybernetics:
  Systems}, 2021.

\bibitem{feng2020safety}
S.~Feng, Y.~Feng, X.~Yan, S.~Shen, S.~Xu, and H.~X. Liu, ``Safety assessment of
  highly automated driving systems in test tracks: A new framework,''
  \emph{Accident Analysis \& Prevention}, vol. 144, p. 105664, 2020.

\bibitem{stern2018dissipation}
R.~E. Stern, S.~Cui, M.~L. Delle~Monache, R.~Bhadani, M.~Bunting, M.~Churchill,
  N.~Hamilton, H.~Pohlmann, F.~Wu, B.~Piccoli \emph{et~al.}, ``Dissipation of
  stop-and-go waves via control of autonomous vehicles: Field experiments,''
  \emph{Transportation Research Part C: Emerging Technologies}, vol.~89, pp.
  205--221, 2018.

\bibitem{liao2021cooperative}
X.~Liao, Z.~Wang, X.~Zhao, K.~Han, P.~Tiwari, M.~J. Barth, and G.~Wu,
  ``Cooperative ramp merging design and field implementation: A digital twin
  approach based on vehicle-to-cloud communication,'' \emph{IEEE Transactions
  on Intelligent Transportation Systems}, 2021.

\bibitem{hoffmann2004stanford}
G.~Hoffmann, D.~G. Rajnarayan, S.~L. Waslander, D.~Dostal, J.~S. Jang, and
  C.~J. Tomlin, ``The stanford testbed of autonomous rotorcraft for multi agent
  control (starmac),'' in \emph{The 23rd Digital Avionics Systems Conference
  (IEEE Cat. No. 04CH37576)}, vol.~2.\hskip 1em plus 0.5em minus 0.4em\relax
  IEEE, 2004, pp. 12--E.

\bibitem{how2008real}
J.~P. How, B.~Behihke, A.~Frank, D.~Dale, and J.~Vian, ``Real-time indoor
  autonomous vehicle test environment,'' \emph{IEEE Control Systems Magazine},
  vol.~28, no.~2, pp. 51--64, 2008.

\bibitem{paull2017duckietown}
L.~Paull, J.~Tani, H.~Ahn, J.~Alonso-Mora, L.~Carlone, M.~Cap, Y.~F. Chen,
  C.~Choi, J.~Dusek, Y.~Fang \emph{et~al.}, ``Duckietown: an open, inexpensive
  and flexible platform for autonomy education and research,'' in \emph{2017
  IEEE International Conference on Robotics and Automation (ICRA)}.\hskip 1em
  plus 0.5em minus 0.4em\relax IEEE, 2017, pp. 1497--1504.

\bibitem{stager2018scaled}
A.~Stager, L.~Bhan, A.~Malikopoulos, and L.~Zhao, ``A scaled smart city for
  experimental validation of connected and automated vehicles,''
  \emph{IFAC-PapersOnLine}, vol.~51, no.~9, pp. 130--135, 2018.

\bibitem{karaman2017project}
S.~Karaman, A.~Anders, M.~Boulet, J.~Connor, K.~Gregson, W.~Guerra, O.~Guldner,
  M.~Mohamoud, B.~Plancher, R.~Shin \emph{et~al.}, ``Project-based,
  collaborative, algorithmic robotics for high school students: Programming
  self-driving race cars at mit,'' in \emph{2017 IEEE integrated STEM education
  conference (ISEC)}.\hskip 1em plus 0.5em minus 0.4em\relax IEEE, 2017, pp.
  195--203.

\bibitem{goldfain2019autorally}
B.~Goldfain, P.~Drews, C.~You, M.~Barulic, O.~Velev, P.~Tsiotras, and J.~M.
  Rehg, ``Autorally: An open platform for aggressive autonomous driving,''
  \emph{IEEE Control Systems Magazine}, vol.~39, no.~1, pp. 26--55, 2019.

\bibitem{o2019f1}
M.~O'Kelly, V.~Sukhil, H.~Abbas, J.~Harkins, C.~Kao, Y.~V. Pant, R.~Mangharam,
  D.~Agarwal, M.~Behl, P.~Burgio \emph{et~al.}, ``F1/10: An open-source
  autonomous cyber-physical platform,'' \emph{arXiv preprint arXiv:1901.08567},
  2019.

\bibitem{keivan2013realtime}
N.~Keivan and G.~Sibley, ``Realtime simulation-in-the-loop control for agile
  ground vehicles,'' in \emph{Conference Towards Autonomous Robotic
  Systems}.\hskip 1em plus 0.5em minus 0.4em\relax Springer, 2013, pp.
  276--287.

\bibitem{gonzales2016autonomous}
J.~Gonzales, F.~Zhang, K.~Li, and F.~Borrelli, ``Autonomous drifting with
  onboard sensors,'' in \emph{Advanced Vehicle Control: Proceedings of the 13th
  International Symposium on Advanced Vehicle Control (AVEC16)}, 2016, p. 133.

\bibitem{hyldmar2019fleet}
N.~Hyldmar, Y.~He, and A.~Prorok, ``A fleet of miniature cars for experiments
  in cooperative driving,'' in \emph{2019 International Conference on Robotics
  and Automation (ICRA)}.\hskip 1em plus 0.5em minus 0.4em\relax IEEE, 2019,
  pp. 3238--3244.

\bibitem{fok2012platform}
C.-L. Fok, M.~Hanna, S.~Gee, T.-C. Au, P.~Stone, C.~Julien, and S.~Vishwanath,
  ``A platform for evaluating autonomous intersection management policies,'' in
  \emph{2012 IEEE/ACM Third International Conference on Cyber-Physical
  Systems}.\hskip 1em plus 0.5em minus 0.4em\relax IEEE, 2012, pp. 87--96.

\bibitem{wan2011general}
J.~Wan, H.~Suo, H.~Yan, and J.~Liu, ``A general test platform for
  cyber-physical systems: unmanned vehicle with wireless sensor network
  navigation,'' \emph{Procedia Engineering}, vol.~24, pp. 123--127, 2011.

\bibitem{sasaki2016vehicle}
K.~Sasaki, N.~Suzuki, S.~Makido, and A.~Nakao, ``Vehicle control system
  coordinated between cloud and mobile edge computing,'' in \emph{2016 55th
  Annual Conference of the Society of Instrument and Control Engineers of Japan
  (SICE)}.\hskip 1em plus 0.5em minus 0.4em\relax IEEE, 2016, pp. 1122--1127.

\bibitem{zheng2020smoothing}
Y.~Zheng, J.~Wang, and K.~Li, ``Smoothing traffic flow via control of
  autonomous vehicles,'' \emph{IEEE Internet of Things Journal}, vol.~7, no.~5,
  pp. 3882--3896, 2020.

\bibitem{la2012development}
H.~M. La, R.~S. Lim, J.~Du, S.~Zhang, G.~Yan, and W.~Sheng, ``Development of a
  small-scale research platform for intelligent transportation systems,''
  \emph{IEEE Transactions on Intelligent Transportation Systems}, vol.~13,
  no.~4, pp. 1753--1762, 2012.

\bibitem{zayas2021digital}
R.~M. Zayas, L.~E. Beaver, B.~Chalaki, H.~Bang, and A.~A. Malikopoulos, ``A
  digital smart city for emerging mobility systems,'' \emph{arXiv preprint
  arXiv:2109.02811}, 2021.

\bibitem{grieves2014digital}
M.~Grieves, ``Digital twin: manufacturing excellence through virtual factory
  replication,'' \emph{White paper}, vol.~1, pp. 1--7, 2014.

\bibitem{rasheed2020digital}
A.~Rasheed, O.~San, and T.~Kvamsdal, ``Digital twin: Values, challenges and
  enablers from a modeling perspective,'' \emph{Ieee Access}, vol.~8, pp.
  21\,980--22\,012, 2020.

\bibitem{tao2018digital}
F.~Tao, H.~Zhang, A.~Liu, and A.~Y. Nee, ``Digital twin in industry:
  State-of-the-art,'' \emph{IEEE Transactions on Industrial Informatics},
  vol.~15, no.~4, pp. 2405--2415, 2018.

\bibitem{wang2021digital2}
Z.~Wang, K.~Han, and P.~Tiwari, ``Digital twin simulation of connected and
  automated vehicles with the unity game engine,'' in \emph{2021 IEEE 1st
  International Conference on Digital Twins and Parallel Intelligence
  (DTPI)}.\hskip 1em plus 0.5em minus 0.4em\relax IEEE, 2021, pp. 1--4.

\bibitem{wang2021digital}
------, ``Digital twin-assisted cooperative driving at non-signalized
  intersections,'' \emph{IEEE Transactions on Intelligent Vehicles}, 2021.

\bibitem{wang2020digital}
Z.~Wang, X.~Liao, X.~Zhao, K.~Han, P.~Tiwari, M.~J. Barth, and G.~Wu, ``A
  digital twin paradigm: Vehicle-to-cloud based advanced driver assistance
  systems,'' in \emph{2020 IEEE 91st Vehicular Technology Conference
  (VTC2020-Spring)}.\hskip 1em plus 0.5em minus 0.4em\relax IEEE, 2020, pp.
  1--6.

\bibitem{kumar2018novel}
S.~A. Kumar, R.~Madhumathi, P.~R. Chelliah, L.~Tao, and S.~Wang, ``A novel
  digital twin-centric approach for driver intention prediction and traffic
  congestion avoidance,'' \emph{Journal of Reliable Intelligent Environments},
  vol.~4, no.~4, pp. 199--209, 2018.

\bibitem{glaessgen2012digital}
E.~Glaessgen and D.~Stargel, ``The digital twin paradigm for future nasa and us
  air force vehicles,'' in \emph{53rd AIAA/ASME/ASCE/AHS/ASC structures,
  structural dynamics and materials conference 20th AIAA/ASME/AHS adaptive
  structures conference 14th AIAA}, 2012, p. 1818.

\bibitem{grieves2017digital}
M.~Grieves and J.~Vickers, ``Digital twin: Mitigating unpredictable,
  undesirable emergent behavior in complex systems,'' in
  \emph{Transdisciplinary perspectives on complex systems}.\hskip 1em plus
  0.5em minus 0.4em\relax Springer, 2017, pp. 85--113.

\bibitem{tepper2017mixed}
O.~M. Tepper, H.~L. Rudy, A.~Lefkowitz, K.~A. Weimer, S.~M. Marks, C.~S. Stern,
  and E.~S. Garfein, ``Mixed reality with hololens: where virtual reality meets
  augmented reality in the operating room,'' \emph{Plastic and reconstructive
  surgery}, vol. 140, no.~5, pp. 1066--1070, 2017.

\bibitem{ohta2014mixed}
Y.~Ohta and H.~Tamura, \emph{Mixed reality: merging real and virtual
  worlds}.\hskip 1em plus 0.5em minus 0.4em\relax Springer Publishing Company,
  Incorporated, 2014.

\bibitem{flavian2019impact}
C.~Flavi{\'a}n, S.~Ib{\'a}{\~n}ez-S{\'a}nchez, and C.~Or{\'u}s, ``The impact of
  virtual, augmented and mixed reality technologies on the customer
  experience,'' \emph{Journal of business research}, vol. 100, pp. 547--560,
  2019.

\bibitem{wang2020augmented}
Z.~Wang, K.~Han, and P.~Tiwari, ``Augmented reality-based advanced
  driver-assistance system for connected vehicles,'' in \emph{2020 IEEE
  International Conference on Systems, Man, and Cybernetics (SMC)}.\hskip 1em
  plus 0.5em minus 0.4em\relax IEEE, 2020, pp. 752--759.

\bibitem{szalai2020mixed}
M.~Szalai, B.~Varga, T.~Tettamanti, and V.~Tihanyi, ``Mixed reality test
  environment for autonomous cars using unity 3d and sumo,'' in \emph{2020 IEEE
  18th World Symposium on Applied Machine Intelligence and Informatics
  (SAMI)}.\hskip 1em plus 0.5em minus 0.4em\relax IEEE, 2020, pp. 73--78.

\bibitem{yang2021multi}
C.~Yang, J.~Dong, Q.~Xu, M.~Cai, H.~Qin, J.~Wang, and K.~Li, ``Multi-vehicle
  experiment platform: A digital twin realization method,'' in \emph{2022
  IEEE/SICE International Symposium on System Integration (SII)}, 2022, pp.
  705--711.

\bibitem{wang2022experimental}
J.~Wang, Y.~Zheng, J.~Dong, C.~Chen, M.~Cai, K.~Li, and Q.~Xu, ``Implementation
  and experimental validation of data-driven predictive control for dissipating
  stop-and-go waves in mixed traffic,'' \emph{arXiv preprint arXiv:2204.03747},
  2022.

\bibitem{cai2021experimental}
M.~Cai, Q.~Xu, C.~Yang, J.~Dong, C.~Chen, J.~Wang, J.~Wang, and K.~Li,
  ``Experimental validation of multi-lane formation control for connected and
  automated vehicles in multiple scenarios,'' \emph{arXiv preprint
  arXiv:2112.00312}, 2021.

\bibitem{hu2015mobile}
Y.~C. Hu, M.~Patel, D.~Sabella, N.~Sprecher, and V.~Young, ``Mobile edge
  computing—a key technology towards 5g,'' \emph{ETSI white paper}, vol.~11,
  no.~11, pp. 1--16, 2015.

\bibitem{xu2017internet}
W.~Xu, H.~Zhou, N.~Cheng, F.~Lyu, W.~Shi, J.~Chen, and X.~Shen, ``Internet of
  vehicles in big data era,'' \emph{IEEE/CAA Journal of Automatica Sinica},
  vol.~5, no.~1, pp. 19--35, 2017.

\bibitem{sasaki2018vehicle}
K.~Sasaki, S.~Makido, and A.~Nakao, ``Vehicle control system for cooperative
  driving coordinated multi-layered edge servers,'' in \emph{2018 IEEE 7th
  International Conference on Cloud Networking (CloudNet)}.\hskip 1em plus
  0.5em minus 0.4em\relax IEEE, 2018, pp. 1--7.

\bibitem{shi2016promise}
W.~Shi and S.~Dustdar, ``The promise of edge computing,'' \emph{Computer},
  vol.~49, no.~5, pp. 78--81, 2016.

\bibitem{wang2022mobility}
Z.~Wang, R.~Gupta, K.~Han, H.~Wang, A.~Ganlath, N.~Ammar, and P.~Tiwari,
  ``Mobility digital twin: Concept, architecture, case study, and future
  challenges,'' \emph{IEEE Internet of Things Journal}, 2022.

\bibitem{wang2021leading}
J.~Wang, Y.~Zheng, C.~Chen, Q.~Xu, and K.~Li, ``Leading cruise control in mixed
  traffic flow: System modeling, controllability, and string stability,''
  \emph{IEEE Transactions on Intelligent Transportation Systems}, vol.~23,
  no.~8, pp. 12\,861--12\,876, 2022.

\bibitem{amer2017modelling}
N.~H. Amer, H.~Zamzuri, K.~Hudha, and Z.~A. Kadir, ``Modelling and control
  strategies in path tracking control for autonomous ground vehicles: a review
  of state of the art and challenges,'' \emph{J. Intell. Rob. Syst.}, vol.~86,
  no.~2, pp. 225--254, 2017.

\bibitem{milanes2013cooperative}
V.~Milan{\'e}s, S.~E. Shladover, J.~Spring, C.~Nowakowski, H.~Kawazoe, and
  M.~Nakamura, ``Cooperative adaptive cruise control in real traffic
  situations,'' \emph{IEEE Transactions on intelligent transportation systems},
  vol.~15, no.~1, pp. 296--305, 2013.

\end{thebibliography}

\end{document}